\documentclass[times,twocolumn,final,authoryear]{elsarticle}

\usepackage{prletters}
\usepackage{framed,multirow}

\usepackage{amssymb}
\usepackage{latexsym}
 \usepackage{verbatim} % comentarios
\usepackage{cite}
\usepackage{amsmath,amsfonts}
\usepackage{algorithmic}
\usepackage{graphicx}
\usepackage{textcomp}
\usepackage{xcolor}
\usepackage{epsfig}
\usepackage{tabu}
\usepackage{multirow} 
\usepackage{adjustbox}
\usepackage{multirow}
\usepackage{array} 
\usepackage{todonotes}
\usepackage{float}
\usepackage[makeroom]{cancel}
\usepackage{rotating}
\usepackage{graphicx,multirow}
\usepackage{lipsum}% http://ctan.org/pkg/lipsum

\usepackage{filecontents}
\usepackage{subcaption}
\usepackage{multicol}% http://ctan.org/pkg/multicols
\usepackage{array,booktabs}
\usepackage{bm}

\usepackage{url}
\definecolor{newcolor}{rgb}{.8,.349,.1}

\journal{Pattern Recognition Letters}

\begin{document}

\ifpreprint
  \setcounter{page}{1}
\else
  \setcounter{page}{1}
\fi

\begin{frontmatter}

\title{Saliency for Free: Saliency Prediction as a Side-Effect of Object Recognition}
%Learning Saliency Prediction from Object Recognition
%Saliency for Free: Saliency Prediction as a Side-Effect of Object Recognition
% by-product side-effect
% Saliency as a Side-Effect of Object Recognition
% Saliency as a By-Product of Object Recognition
%Saliency Prediction Unsupervised
% Saliency Emerging from Object Recognition

\author[1,2]{Carola \snm{Figueroa-Flores}\corref{cor1}} 
\cortext[cor1]{Corresponding author: }
\ead{cafigueroa@cvc.uab.es}
\author[1]{David \snm{Berga}}
\author[1]{Joost \snm{van de Weijer}}
\author[1]{Bogdan \snm{Raducanu}}

\address[1]{Computer Vision Center, Edifici ``O'' - Campus UAB, 8193 Bellaterra (Barcelona), Spain}
\address[2]{Department of Computer Science and  Information Technology, Universidad del B\'io  B\'io, Chile}

%\received{1 May 2013}
%\finalform{10 May 2013}
%\accepted{13 May 2013}
%\availableonline{15 May 2013}
%\communicated{S. Sarkar}

\begin{abstract}
Saliency is the perceptual capacity of our visual system to focus our attention (i.e. gaze) on relevant objects. Neural networks for saliency estimation require ground truth saliency maps for training which are usually achieved via eyetracking experiments. 
In the current paper, we demonstrate that saliency maps can be generated as a side-effect of training an object recognition deep neural network that is endowed with a saliency branch. Such a network does not require any ground-truth saliency maps for training.
%In the current paper, \BR{based on a deep learning framework, we demonstrate that it is no longer necessary to provide saliency maps as ground-truth for training, instead they are being learnt} as a side-effect of training an object recognition network that is endowed with a saliency branch. 
Extensive experiments carried out on both real and synthetic saliency datasets demonstrate that our approach is able to generate accurate saliency maps, achieving competitive results on both synthetic and real datasets when compared to methods that do require ground truth data. 

\end{abstract}


\end{frontmatter}

%\linenumbers

\section{Introduction}
\label{sec1}
%\alertJW{We have 7p}
One of the perceptual cues used for scene understanding is image saliency, i.e. a representation of the scene that highlights those regions which are more informative than their surroundings. Thus, by selecting regions which appear relevant based on saliency maps, we could discard the rest of of the image (usually the background). Therefore, saliency detection could be considered a valuable pre-processing step for a wide range of applications. For example, it has been successfully applied for facial features detection and localization \citep{Jian2014facialfeature}, increasing the local contrast for underwater imaging \citep{Jian2018underwater}, modeling spatiotemporal saliency in videos \citep{Liu2017videosaliency,Zhou2018videosaliency}, and modeling the atypical visual attention in children with autism spectrum disorder (ASD) \citep{Wei2019ASD,Wei2020ASD}, to name just a few. In a different direction, some other approaches addressed the problem of salient object detection using a visual-attention-aware model \citep{Jian2015patches} or by fusing the high-level RGB and depth features in an interactive and adaptive way \citep{Li2020ICNet}.

Computational methods in saliency detection used in computer vision are intended to determine which regions of the image attract humans' attention. Saliency methods can be divided in two main categories: (i) salient object detection methods (which segment relevant objects in the image) \citep{Zhang2018cvpr,Zhao2019iccv}; and (ii) methods which produce eye-fixation maps \citep{huang2015salicon,Pan2016,murabito2018top}. For the second category, which is the focus of this article, the common way to obtain an accurate saliency map is to perform eye tracking experiments on still images. Eye fixations from different participants are fused to obtain a unique map, named fixation map, which will represent the saliency ground truth. These (binary) fixation maps are then smoothed by 1 degree of visual angle (dva or deg) in order to simulate the average deviation of capture of the eye tracker \citep{LeMeur2012,Torralba2006}. This smoothing is usually done using a circular gaussian filter, obtaining a continuous representation of the saliency map. The saliency map is assumed to be specific for each image (depending on image features), but experimentation may induce certain patterns such as the center bias.  The center bias (CB) is the common region where participants tend to look, this can be due to:  (i) photographs tend to frame the salient object centered on the imaage, (ii) there are oculomotor tendencies from the task focusing the gaze on the center \citep{NAKASHIMA201559} and (iii) some images do not show objects salient enough to focus attention outside the center.  This center bias is present in most saliency datasets and is also exploited by several saliency models to better simulate human data.

%Therefore, we will here look into how to extend our method with the center bias; we will consider botha supervised and unsupervised center bias approach.}
%~\citep{itti1998model}~\citet{itti1998model} 

\begin{table*}[tb]
\centering
\scriptsize
\setlength{\tabcolsep}{5pt}
\caption{Description of saliency models}
\begin{tabular}{ |c|c|c|c|c|c|c| } 
\hline
Name & Year & Features/Architecture & Mechanism & Learning & Training Data ($\#$img) & Bias/Priors\\
\hline
IKN & 1998 & DoG (color+intensity) & C-S & - & - & -\\
AIM & 2005 & ICA (infomax) & max-like & Unsupervised & Corel (3600) & -\\
GBVS & 2006 & Markov chains & graph prob. & Unsupervised & Einhauser (108) & graph norm. \\
SDLF & 2006 & Steerable pyramid & local+global prob. & Unsupervised & Oliva (8100) & scene priors \\
\hline
ML-Net & 2016 & VGG-16 & Backprop.(finetuning) & Supervised & SALICON (10k), MIT (1003) & learned priors \\
DeepGazeII & 2016 & VGG-19 & Backprop.(finetuning) & Supervised & SALICON (10k), MIT (1003) & center bias \\
SAM &  2018 & VGG-16/ResNet-50+LSTM & Backprop.(finetuning) & Supervised & SALICON (10k) \& others & gaussian priors\\
SalGAN & 2017 & VGG-16 Autoencoder & Finetuning+GAN Loss & Supervised & SALICON (10k), MIT (1003) & - \\
\hline
\end{tabular}\\
DoG: difference of gaussians, ICA: independent component analysis, C-S: center-surround, max-like: max-likelihood probability, BCE: binary cross-entropy, GAN: Generative adversarial network

\label{table:models}
\end{table*}

~\citet{itti1998model}  proposed  one  of  the  first  computational  saliency  methods  based  on combining the saliency cues for color, orientation and luminance.  Many works followed  proposing  a large  variety  of  hand-crafted  features  for  saliency  \citep{Ramanathan2010,Borji2014}. In the last decade, computational saliency estimation has moved from handcrafted to deep features \citep{Li2016deepsaliency}. These methods aim to find a network that computes saliency maps that are close to ground truth saliency maps. A limitation of these approaches is that they require saliency ground truth for their training. Generating saliency ground truth is a costly process and is required for each new dataset, and affects the efficiency of these approaches.

In the human visual system, saliency is applied to select a small part of the incoming sensory information. As a result, massive sensory input can be processed despite limited computational capacity of the brain~\citep{itti2001computational}. It allows humans to rapidly and efficiently process the incoming information. The capability to attend the most relevant information in the image present in the human visual system could also be important for neural networks that aim to process visual data. In this paper, we endow a neural network that aims to perform object recognition with a separate branch that computes a saliency map. This map is used to attend to specific regions in the image (thereby selecting the part of the information deemed most relevant). The potential of such a network is that it can be trained on any image classification dataset. The saliency maps would be the side-effect of training this network, and hence our method allows for the computation of saliency without needing any eyetracking ground truth data to train the deep neural network. 

%Recently, \citet{Figueroa2019} showed that saliency maps could successfully be used as an attention mechanism, to improve  object classification. They validated their approach on several fine-grained recognition datasets. A limitation of this approach is that saliency ground-truth is required to train a saliency estimation network. Generating saliency ground-truth is a tedious process and is required for each new dataset, and affects the efficiency of this approach.

%To overcome the aforementioned limitation, in this paper we introduce an approach for saliency estimation in an unsupervised way. 

In this paper, we evaluate the accuracy of the saliency maps that are produced as a side-effect of object recognition. Additionally, we also evaluate the usage of supervised and unsupervised CB in our framework. We show that the CB improves in most datasets where the CB is more present. To summarize, our main contributions are:
\begin{itemize}
\item We demonstrate that it is possible to obtain accurate saliency maps by training an object recognition network endowed with a saliency branch. Our method does not require any saliency ground truth data. 
\item We include an extensive study of the effect of center bias on the results.
\item Extensive experiments performed on real and synthetic image datasets show that highly accurate saliency maps are obtained. Our method obtains competitive results on several standard benchmark datasets and the new state-of-the-art on the CAT2000 dataset. 
%This opens the possibility to obtain saliency maps on-the-fly thus eliminating the need for experimentation or annotation, which are very tedious tasks
\end{itemize}

%The paper is structured as follows: Section 2 contains the related work. In Section 3 we present our method for unsupervised saliency and CB estimation. In Section 4 we present the experimental results performed on real and synthetic datasets. Finally, in section 5 we draw our conclusions and present the guidelines for future work.
The current work is related to our earlier work~\citep{Figueroa2020}. There we focus on fine-grained image classification, and show that a saliency branch can be used to improve results. In this paper, we show that a saliency branch trained for image classification can actually obtain competitive results on the saliency benchmark dataset, without requiring any saliency ground truth data for training. To the best of our knowledge, we are the first to show that saliency prediction can be obtained as a side-effect of object recognition.

\section{Related Work}

%The current work is an extension of our earlier work~\citep{Figueroa2020}. Here we extend the work by including a study of the center bias which allows us to further improve the results. In addition, we also extend the analysis to three new datasets (CAT2000, MIT1003 and KTH). 

\subsection{Saliency models} 
\label{sec:saliencymodels}
Initial work on computational saliency was defined by~\citet{itti1998model}, introducing a framework for obtaining a unique saliency map from an image. This work extracts multi-scale and multi-orientation features with DoG filters, aiming to simulate simple cell computations found in the visual cortex in the brain. These maps are fused to a unique saliency map using winner-take-all mechanisms. This framework has inspired several models~\citep{Borji2013c,Zhang2013,Riche2016a}, mainly varying on the feature extraction part (either handcrafted or trained). For instance, the unsupervised model AIM~\citep{Bruce2005} uses a dictionary of images in order to train sparse priors. These priors are learned and then computed with the feature extractor filters. Later,~\citet{Bruce2005} combine IKN feature extraction with AIM's information maximization and then modulate the resulting regions to psychophysical data. Similarly, \citet{Torralba2006} proposes a contextually-modulated saliency model (SDLF) which is based on task priors when observing real scenes. Latest models (e.g. ML-Net \citep{mlnet2016}, SAM \citep{Cornia2018tip}, DeepGazeII \citep{Kummerer2016}, SalGAN \citep{Pan2017SalGAN2}) use fixation data from image saliency datasets (i.e. that provide eye tracking data) as ground truth for learning the saliency map with CNN architectures. These models usually train a neural network that focuses on the most salient regions of the input image to iteratively refine the predicted saliency map (see more details in Table~\ref{table:models}). ML-Net learns a prior map based on the common ground truth saliency maps, acting as a mask. This is multiplied by the output map of the network on training saliency. In DeepGazeII, they sum a probability distribution (baseline of fixations) over the image. Instead, SAM utilizes an LSTM and trains a set of Gaussian parameters acting as an attentive mechanism to the final map, which is finetuned with human fixation density maps. Finally, SalGAN uses an autoencoder architecture, which is trained with prediction in combination with an adversarial loss.  

\subsection{Center bias}
Eye movement datasets used for saliency evaluation tend to be center biased (most fixations tend to be at the center of the image). Several factors on the experimentation and the stimuli can cause this effect. For instance, most real images frame the scene (the relevant or salient part is in the center of view in photographies). Non-salient/non-popout stimuli \citep{Tatler2007,tatler2008,Berga2019a} has been shown to promote center biases, as participants do not have any region to attend to, specially if the task sometimes involves centering the gaze on the image. These center biases have an influence on how to evaluate saliency models upon predicting fixations \citep{Bylinskii2016,Borji2016}, as these fixations are accounted while are not specific to image saliency. Some other works used CB in order to improve saliency detection \citep{Jian2018patches,Jian2021positionprior}.

\subsection{Saliency features for image classification}
The vast majority of saliency methods previously reviewed are evaluated on the task of how accurate their generated saliency maps are. Therefore, it is questionable whether training other tasks (such as image classification) can also represent saliency as opposed to uniquely training on biased fixation data (which is distinct for every dataset or experimentation setting). The question of whether saliency is important for object recognition and object tracking has been raised in \citep{vasconcelos2010saliency}. This is also the purpose of \citep{Figueroa2019}, where the authors investigate to what extent saliency information can be exploited to improve object recognition when the available training data is scarce. The authors designed a two-branch image classification deep network, where one of the branches takes saliency information as input.
The network processes the saliency through the dedicated branch and uses the resulting saliency features to modulate the visual features from the standard RGB branch, thus forcing the upper layers to focus on the relevant parts only. In the same line, \citep{murabito2018top} learned to generate saliency maps from RGB images, but in this case their method is supervised. However, none of these methods shows that saliency maps can be computed as a side-effect of an end-to-end trained object recognition network. 

%In prior work, we showed that saliency maps could successfully be used as an attention mechanism to improve object classification~\citet{Figueroa2019}. We validated this approach on several fine-grained recognition datasets. A limitation of this approach was that saliency ground-truth is required to train a saliency estimation network. In this paper, we show that no saliency maps are required to train the network, and that these can be computed by end-to-end training of an object recognition network with saliency branch. 

\begin{figure}[!h]
\begin{center}
\includegraphics[width=0.49\textwidth]{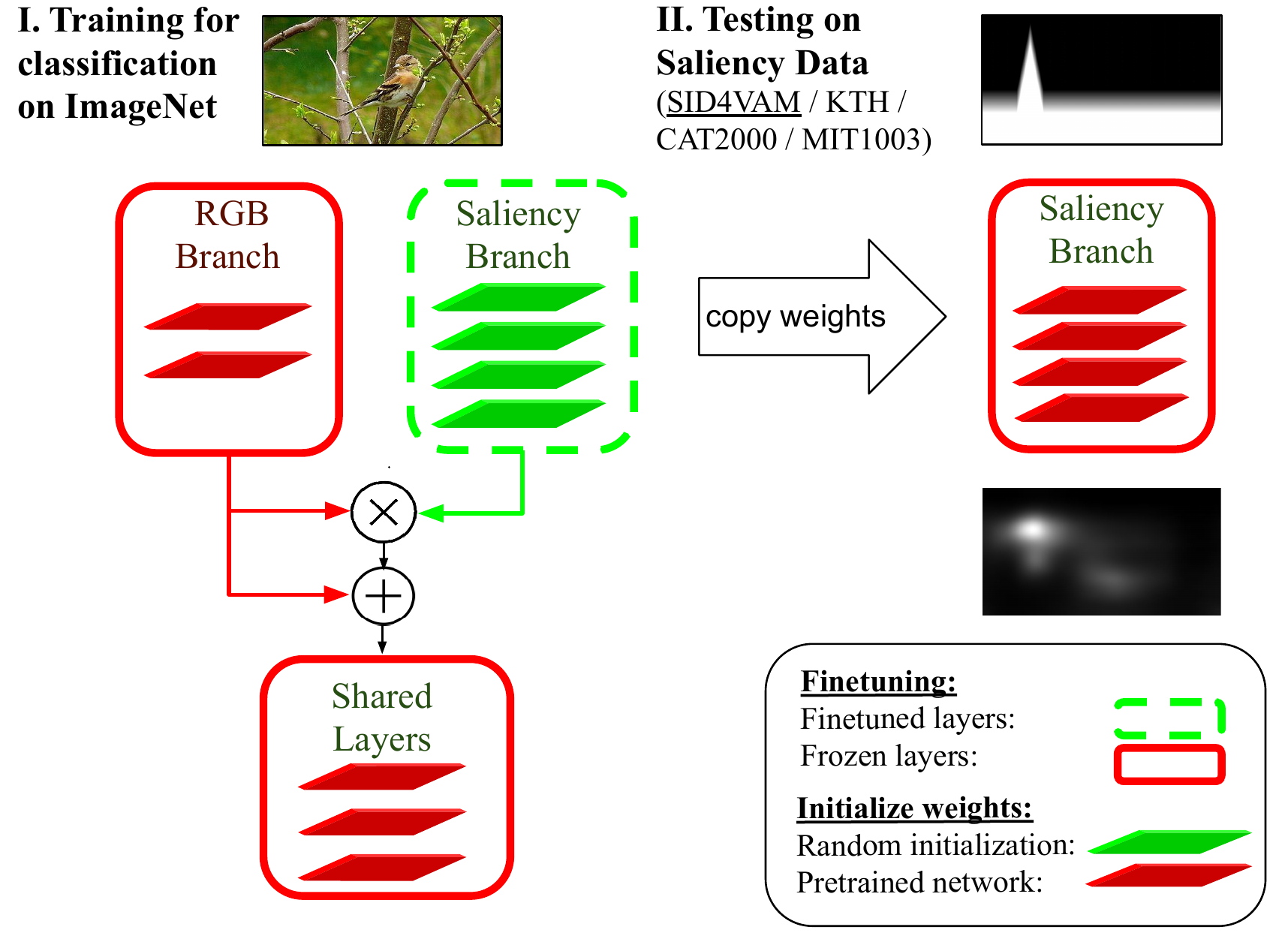}
\end{center}
\caption{Overview of our method. We process an RGB input image through two branches: one branch extracts the RGB features and the other one is used to learn saliency maps.}%\alertJW{If we need space we could consider this one column version} } 
\label{fig:overview}
\end{figure}

\section{Proposed Method}
\subsection{Network architecture}
%\subsection{Overview of the Method}
The overview of our proposed method is shown in Figure~\ref{fig:overview}. %, which is based on the work presented in \citep{Figueroa2020}.
The network consists of two branches: one to extract the features from an RGB image (the red branch called \emph{RGB branch}), and the other one (called the \emph{saliency branch} marked in green) to generate the saliency maps from the same RGB image. 
Both branches are combined using a \emph {modulation layer} (represented by the $\otimes$ symbol) and the output is further processed by several shared layers ending up with a classification layer.

%\alertJW{Here should be details on the modulation layer. Maybe equation. It would be nice if this does not refer to any fixed dimensionalities of a particular architecture. }

Consider an input image $ I (x_1, x_2, x_3) $, where $x_1, x_2$ are the spatial coordinates and $ x_3 = \{1,2,3\} $ indicate the three color channels of the image. Let us define the three networks as being $s$ for the saliency branch, $r$ for the RGB branch and $f$ for the final shared layers. We will name the output of the saliency branch the saliency image $S (x_1, x_2 )$ (we will design the saliency branch to output only a single saliency image, therefore there are only two coordinates involved), and  the output of the RGB branch $R (x_1, x_2, x_3)$. Both $S$ and $R$ will have the same spatial resolution. We now define the modulation layer as: \begin{equation} 
\begin{split}
\mathring R  \left( x_1,x_2,x_3 \right) &= r \left( I \left(x_1,x_2,x_3\right) \right) \cdot \left( s\left( I\left(x_1,x_2,x_3\right) \right) + 1  \right)  \\
& = R \left( x_1,x_2,x_3 \right) \cdot  S\left( x_1,x_2 \right) + R \left( x_1,x_2,x_3 \right) \label{eq:forward2}.
\end{split}
\end{equation}
Note that the same saliency branch output $S$ is applied to all the feature maps of $R$ (along the $x_3$ dimension). The output $\mathring R$ is a summation of the modulated output  $R \cdot S$ and a non-modulated version of the RGB branch $R$ (see also the skip connection represented by $\oplus$ in Figure~\ref{fig:overview}). This was found to improve results in~\citep{Figueroa2019}. The output of the modulation layer is then used as an input to the shared layers to obtain the final prediction over the classes $y$:
\begin{equation} 
p \left(y|I \right) = f \left( \mathring R \right)\label{eq:netowrk_output},
\end{equation}
where we omit the spatial coordinates for clarity. We train the network for the task of image classification on a training dataset $\mathcal{D}$ of images with the cross-entropy loss: 
\begin{equation}
    \mathcal{L}=\sum_{I\in \mathcal{D}} log  p_{ {c(I)}}\left(y|I \right)     \label{eq:loss},
\end{equation}
where $\mathcal{D}$ is the entire training dataset and $c(I)$ is the ground truth label of image $I$ and $p_c$ is the $c$-th element of the vector $p$. 
% Also consider a saliency map  $\mathring s$ is trained end-to-end and estimated from the same input image $I$. Then, the combination of saliency map $\mathring s$ , which is the output of the saliency branch, and the RGB branch is done with modulation. Consider the output of the $ i ^ {th} $ layer of the network, $ l ^ i $, with dimension $w_i \times h_i \times z_i $. Then we define the modulation as: 
% \begin{equation}
% \hat l^i \left( {x,y,z} \right) = l^i \left( {x,y,z} \right) \cdot \mathring s\left( {x,y}\right),  \label{eq:forward}
% \end{equation}

% In addition to the formula in Eq.~\eqref{eq:forward} we also introduce a skip connection from the RGB branch to the beginning of the joint branch, defined as: 
% \begin{equation}
% \hat l^i \left( {x,y,z} \right) = l^i \left( {x,y,z} \right) \cdot \left( \mathring s\left( {x,y} \right) + 1 \right)\label{eq:forward2}.
% \end{equation}

The RGB branch followed by the modulation layers resembles a standard image classification network (see layers marked in red in the Figure~\ref{fig:overview}-left). In this work, we will consider several architectures, including AlexNet \citep{krizhevsky2012imagenet}, VGG16 \citep{Simonyan15}, and ResNet152 \citep{Resnet50}. The saliency branch consists of four convolutional layers, similar to the first three layers of the AlexNet architecture combined with a 1x1 convolutional layer. More precisely, the output of the third convolutional layer, i.e. the one with 384 dimensional feature maps with a spatial resolution of 13 × 13 (for a 227 × 227 RGB input image), is further processed using a 1 × 1 convolution and then a ReLU activation function. This 1x1 convolution maps the feature map to a single output feature map, and its goal is to calculate the score for each "pixel" and to produce a single map that can be used to modulate the RGB branch. Finally, to generate the input for the posterior classification network, the 13 × 13 saliency maps are upsampled at 27 × 27 (which is the default input size of the following classification module) through bilinear interpolation.

What differentiates our architecture from a standard object recognition network, is the introduction of the saliency branch which transforms the RGB input image into a \emph{modulation map} $S$. While training the network the modulation map learns to focus on those features that are important to perform the classification task. This is a very similar task as for which the human visual system is thought to use visual saliency, namely to identify those regions of high information in the image. In this paper, we show that this modulation map resembles a saliency map. Actually when compared to saliency maps obtained from human eye-tracking experiments, this modulation map is found to provide a surprisingly good estimate of them.   

\subsection{Training the saliency branch}
Our approach is depicted in Figure~\ref{fig:overview}. The main idea is to train the saliency branch $S$ on a classification task. By optimizing the network to be good in image classification, we hypothesize that the saliency branch will learn a mapping from the image $I$ to something similar as a saliency map. The modulation map $S$ will provide higher values to those regions that are important  to performing the image classification task. The learned network $s$ will then be evaluated on several existing saliency estimation datasets. Interestingly, the network $s$ has not been trained with any saliency ground truth, rather the saliency network is trained as a side-effect of training a network optimal for object recognition. 

We would like the classification task to be very general to ensure that the saliency network is trained on a wide variety of images. We therefore choose to train the network on the ImageNet dataset~\citep{krizhevsky2012imagenet} which has 1000 different classes, including classes from plants, sports, artefacts, animals, etc.

As explained above, the purpose of the saliency branch is to generate a saliency map directly from an RGB input image. This network is built by initializing the RGB branch with ImageNet pre-trained weights. The weights of the saliency branch are initialized randomly using the Xavier method \citep{Xavier} (see Figure~\ref{fig:overview}, green layers). Then, the network is selectively trained: we allow to train only the layers corresponding to the Saliency branch (represented by the surrounding green dotted line box) and to freeze all the remaining layers (represented by the solid red line boxes). During training, the saliency branch learns to focus on those regions of the image that are important for the classification of the 1000 ImageNet classes.

Once the Imagenet training is finished, we only use the saliency branch, freeze its weights, and test it on the images of various saliency estimation datasets (see Figure~\ref{fig:overview}-right). We will consider both datasets with real images (Toronto \citep{Bruce2005}, MIT1003 \citep{Judd2009}, KTH \citep{Kootstra2011}) as well as datasets that contain synthetic images (CAT2000 \citep{CAT2000} and SID4VAM \citep{Berga2019ICCV}).

\begin{table}[t]
\scriptsize
\centering
\caption{Simulating the Center Bias by parametrizing Gaussian.}
\begin{tabular}{ |c|c|c| } 
\hline
DVA & Circular & Ellipsoid\\
\hline
36 x 2 & \includegraphics[width=0.9in,height=0.44in]{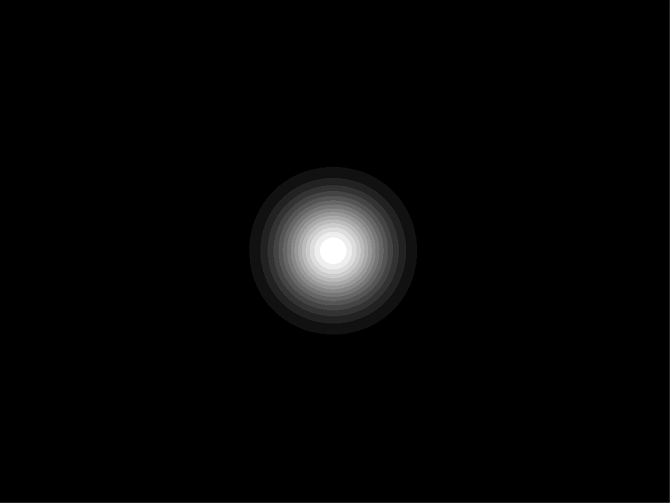} & \includegraphics[width=0.9in,height=0.44in]{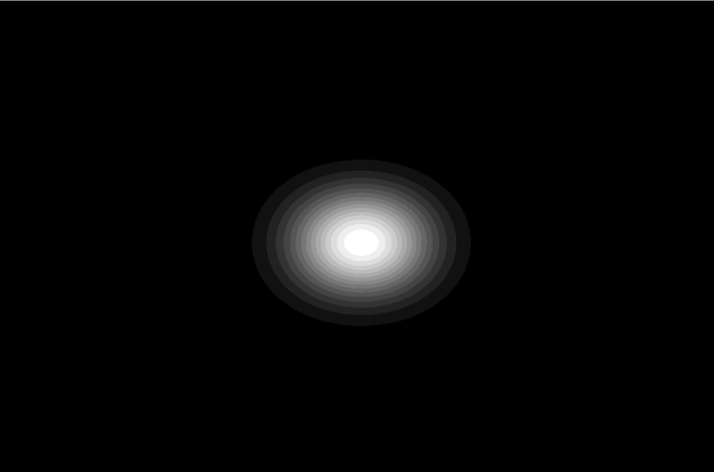}\\ 
36 x 5 &  \includegraphics[width=0.9in,height=0.44in]{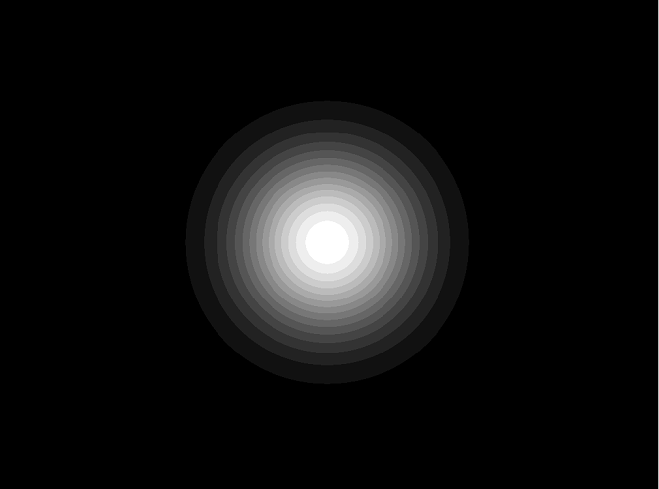} & \includegraphics[width=0.9in,height=0.44in]{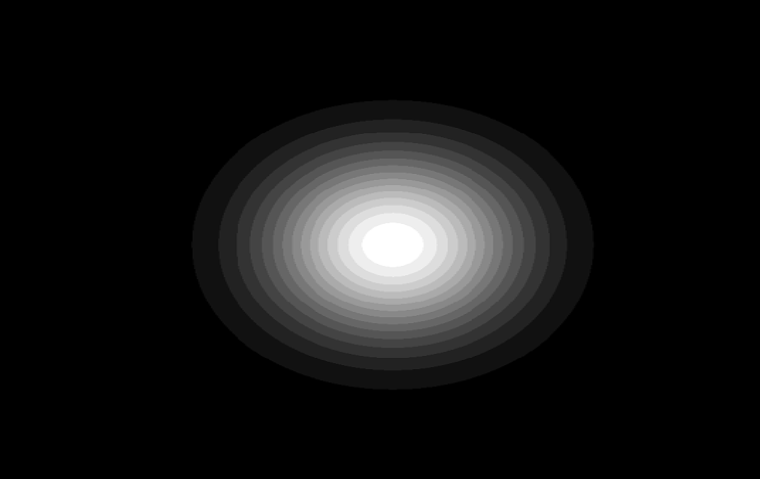}\\
36 x 14 &  \includegraphics[width=0.9in,height=0.44in]{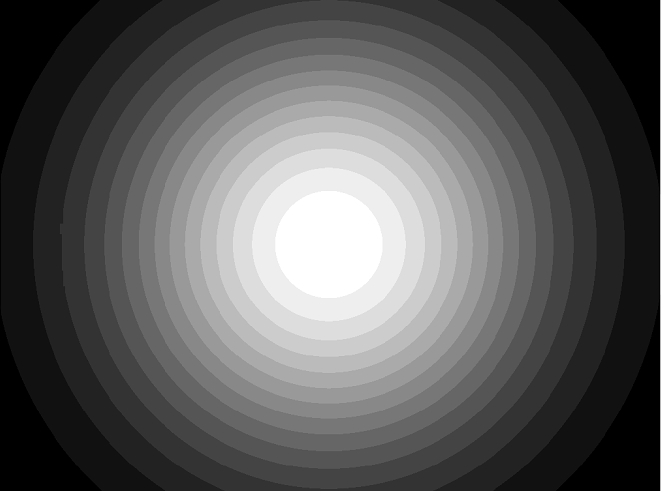} & \includegraphics[width=0.9in,height=0.44in]{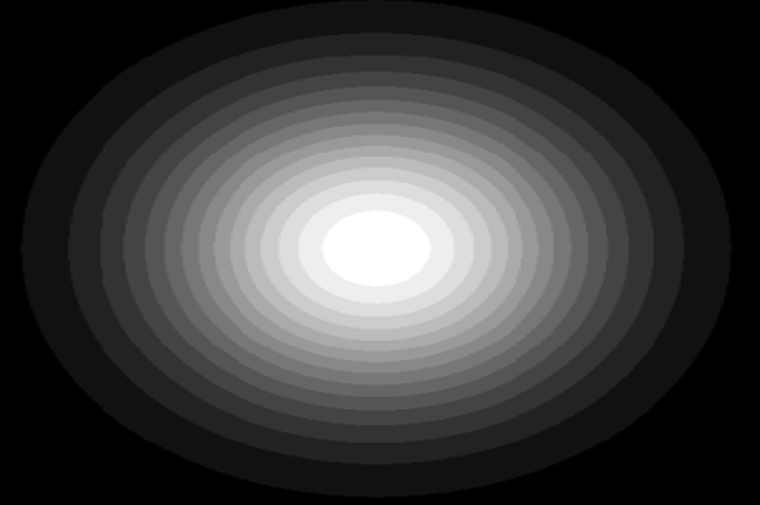}/\\ 
\hline
\end{tabular}
\label{table:centerBias}
\end{table}

\subsection{Combination with center bias} % or Single Gaussian}

%The ground truth of saliency is obtained by locating fixations in the scene. These (binary) fixation maps are then smoothed by 1 degree of visual angle (dva or deg) in order to simulate the average deviation of capture of the eye tracker \citep{LeMeur2012,Torralba2006}. This smoothing is usually done using a circular gaussian filter, obtaining a continuous representation of the saliency map. Several image processing and computer vision techniques have been used in order to accurately represent saliency maps {(see Section \ref{sec:saliencymodels})}.
%The saliency map is assumed to be specific for each image (depending on image features), but experimentation may induce certain patterns such as the center bias. The center bias (CB) is the common region where participants tend to look, this can be due to: (i) photographs tend to frame the salient object centered on the image, (ii) there are oculomotor tendencies from the task focusing the gaze on the center  \citep{NAKASHIMA201559} and (iii) some images do not show objects salient enough to focus attention outside the center.

As we have already mentioned in the introduction, center bias~(CB) is present in most saliency datasets and is also exploited by several saliency models to better simulate human data. Therefore, we will here look into how to extend our method with the center bias; we will consider both a supervised and unsupervised center bias approach.

\paragraph{Supervised CB (SCB)} %The Gaussian function can compute the exact center bias from data with 1 deg of GVA and overlapping all binary fixation maps (seemingly obtaining a unique fixation density map with images put altogether). 
To compute the center bias map, we split the data in two sets, generating the center bias for each of them and evaluating each sample with the opposite split. 

\paragraph{Unsupervised CB (UCB)} To compute the center bias~(CB), we use a 2D Gaussian low-pass filter with $\sigma$=DVA$/(2\sqrt{2 log 2})$, with a window of 6$\sigma$ x 6$\sigma$. Using a parameter "DVA" (degree of visual angle) as a multiplying factor of the pixels. This is the usual smoothing function for building the fixation density maps~\citep{mit-saliency-benchmark,Bruce2015}.

For the UCB we used circular and ellipsoidal versions of the Gaussian function. We did this as center biases might vary on the display and experimental methods for each saliency dataset. %, in most cases these maps are spread in the horizontal axis, as images are usually of rectangular shape.
For the ellipsoid case, we resized the image so that the resulting map is stretched horizontally with a factor of +50\%, but keeping the same DVA vertically.

We selected the DVA according to the following rule: 2 deg corresponds to the approximate maximum diameter of coordinate deviation permitted during eye tracking calibration, this is approximately two times the common deviation of participant's fixations \citep{LeMeur2012, Torralba2006}, 5 deg corresponds to the degrees of higher visual accuity of foveal/central vision \citep{Strasburger2011} and 14 deg corresponds to the radius of the screen (about 512px). See Table~\ref{table:centerBias} for examples of the used center biases. 

\paragraph{Fusion} 

Previously, other models (see Table \ref{table:models} - column 7) used additional computations from priors or baselines from fixation data. For instance, DeepGazeII summed the center baseline whereas ML-Net and SAM the learned priors are used for modulating the result of the network. We defined two regimes for fusing the RGB and the saliency branch: sum  or multiplication. With this we can test at distinct baselines the effect of the center bias over the saliency map produced by the network. See in Table~\ref{table:fusion} different examples of the resulting fusion (sum or multiplication).

\begin{table}[t]
\scriptsize
\centering
\caption{Characteristics of eye tracking datasets}
\begin{tabular}{ |c|c|c|c|c|c| } 
\hline
Dataset & Type & \# Images & \# PP & pxva & Resolution\\ 
\hline
TORONTO & Indoors \& Outdoors & 120 & 20 & 32 & 681x511\\ 
MIT1003  & Indoors \& Outdoors & 1003 & 15 & 35 & 1024x768\\
%NUSEF \cite{Ramanathan2010} & FV & 758 & 25 &   & \cmark \\
KTH$_n$ & Nature photos & 99 & 31 & 34 & 1024x768\\ 
%MIT300 \cite{Judd2012} & FV & 300 & 39 &    & \cmark \\ 
%CAT2000 \cite{CAT2000} & FV & 4000 & 24 &   & \cmark \\ 
CAT2000$_p$ & Synthetic Patterns & 100 & 18 & 38 & 1920x1080\\ 
SID4VAM & Synthetic Pop-out & 230 & 34 & 40 & 1280x1024\\ 
\hline
\end{tabular}
\\\hspace{2mm}\small pxva: pixels per 1 degree of visual angle, PP: participants
\label{table:datasets}
\end{table}

\begin{table*}[!h]
\scriptsize

\centering
\caption{Benchmark of our method with different networks. Top-1 networks are in bold.}
\begin{tabular}{c|c|cccccc} 
 \hline\hline
  Dataset & Model & AUC-Judd & AUC-Borji &  CC & NSS & KL$\downarrow$ & SIM  \\ \hline
  
    & AlexNet & 0.7679 & 0.7308 &  0.4546 & 1.3718 & \textbf{1.5134} & 0.3944 \\
  TORONTO & VGG16 & 0.7812 & \textbf{0.7475} &  0.4627 & 1.4045 & 1.5179 & 0.4201 \\
      & ResNet152 & \textbf{0.7816} & 0.7323 &  \textbf{0.5378} & \textbf{1.6433} & 1.6991 & \textbf{0.4390} \\\hline
  
         & AlexNet  & 0.7323 & 0.7034 &  \textbf{0.2597}& \textbf{0.8654} & \textbf{1.7622} & 0.2844\\
  MIT1003 & VGG16   & \textbf{0.7402} & \textbf{0.7199} &  0.2594 & 0.8597 & 1.7772& \textbf{0.2899} \\
        & ResNet152 & 0.7231 & 0.7084 &  0.2531 & 0.8550 & 2.0785& 0.2839 \\\hline

      & AlexNet   & 0.5975 & 0.5881 &  0.2249 & 0.3374 & \textbf{1.0083} & \textbf{0.5112} \\
  KTH & VGG16     & 0.6028 & 0.5793 &  0.2250 & 0.3459 & 1.3194 & 0.4848 \\
      & ResNet152 & \textbf{0.6154} & \textbf{0.5869} &  \textbf{0.2942} & \textbf{0.4436} & 1.3492 & 0.4989 \\\hline
  
        & AlexNet & 0.7005 & 0.6710 &  0.2950 & 0.7468 & 1.4615 & 0.3936 \\
  CAT2000 & VGG16 & 0.7113 & 0.6741 &  \textbf{0.3151} & 0.8371 & 1.4510 & 0.4031 \\
      & ResNet152 & \textbf{0.7217} & \textbf{0.6805} &  0.3100 & \textbf{0.8548} & \textbf{1.2876} & \textbf{0.4064} \\\hline
  
        & AlexNet & \textbf{0.7413} & \textbf{0.7216} &  \textbf{0.3889} & \textbf{1.4256} & \textbf{1.6652} & \textbf{0.4085} \\
  SID4VAM & VGG16 & 0.6752 & 0.6506 &  0.2707 & 0.8477 & 1.9129 & 0.3695 \\
      & ResNet152 & 0.6988 & 0.6723 &  0.3010 & 1.1140 & 1.9790 & 0.3786 \\\hline
\end{tabular}
\label{table:databases}

\end{table*}

\section{Experiments}
We have performed the evaluation of our approach on five  current eye movement datasets which provide fixations and scanpaths from real scenes during free-viewing tasks. These datasets are composed of real image scenes  (Toronto, MIT1003), natural scenes (KTH) and synthetic images (CAT2000, SID4VAM). See  Table~\ref{table:datasets} for an overview. 

In order to evaluate how accurate the saliency map is able to match the location of human fixations, we use a set of metrics previously defined by~\citet{Borji2013,Bylinskii2016}. The area under ROC (AUC) considers as true positives the saliency map values that coincide with a fixation and false positives the saliency map that have no fixation, then computes the area under the curve. We have used three metrics based on AUC, namely AUC-Judd, AUC-Borji and shuffled AUC (sAUC). Similarly, the  Normalized Scanpath Saliency (NSS) computes the average normalized saliency map that coincide with fixations. Other metrics such as Correlation Coefficient (CC), Kullback-Leibler divergence (KL), similarity (SIM) compute the score on the region distribution statistics of all pixels (KL calculates the divergence and CC/SIM the histogram intersection or similarity of the distribution).

%$TEXT FROM THESIS DAVID: The shuffled AUC (sAUC) is the metric we used for our psychophysical experimentation. It computes the area under ROC considering TP as fixations inside the saliency map, similarly to the AUC. However, this metric does not evaluate FP at random areas of the image but instead uses fixations inside other random images from the same dataset over several trials (10 by default). This metric gives a more accurate evaluation of predicted maps with respect human fixations but penalizing for higher model center biases (which are or can be present for distinct images in the ground truth).

%In Tables~\ref{table:fusion} and \ref{table:SOA} we show results for our model with different parametization and our best setting in comparison with the state of the art saliency models (using both location- and distribution-based metrics; AUC-Judd and SIM). In Table~\ref{table:databases} we show results of Area Under ROC (AUC),  Correlation Coefficient (CC), Normalized Scanpath Saliency (NSS), Kullback-Leibler divergence (KL), similarity (SIM) for every network for all datasets.

% EXP 1: First experiment: multiple networks
After computing the saliency maps for all datasets  (see Table~\ref{table:databases}) with AlexNet, VGG16 and ResNet152 we observed that metric scores vary considerably depending on the network: AlexNet is shown to provide best results for pop-out patterns (SID4VAM) whereas ResNet152 and VGG16 shows overall higher scores with real images (MIT1003, TORONTO, KTH). %In the remainder of the paper, we perform experiments with ResNet152. 

% EXP 2: ablation CB
%\BR{For the next 3 paragraphs we should be more coherent: talk about tables 7-8, then go back to table 5 and then return to tables 7-8. The description of table 5 should contain a reference to the saliency for multiple objects in the image, as requested by R2} 

We have also performed an ablation study for evaluating the effect of the center bias and the fusion (see  Table~\ref{table:fusion}). We tested the center bias extracted from the data (SCB) as well as our unsupervised implementation (UCB) using circular and ellipsoidal Gaussians, testing both fusions with sum and multiplication. For most datasets, the UCB ellipsoid obtains highest scores using a DVA factor of 14 deg and the Sum fusion. For the cases of SCB, fusion with Mult score higher, although both fusions gave very similar results. In the remaining experiments (in Table~\ref{table:SOA_realImages} and Table~\ref{table:SOA_SyntheticsImages}), we use these settings when applying the center bias. 

% visual results EXP2
%older
%We can see in Table~\ref{table:qualitative} different examples of images and the generated saliency maps from different scenes (one per each dataset), including an illustration of each dataset center bias. We can observe that the fusion is able to modulate the saliency map, showing that it can be better to use one strategy or another.

We can see in Table~\ref{table:qualitative} different examples of images for saliency prediction for real (Toronto, MIT1003) and synthetic images (SID4VAM). Our model performs best on detecting pop-out effects on the synthetic images (from visual attention theory \citet{itti1998model}), whilst performing similarly for Toronto. It is to consider that some deep saliency models use several mechanisms to leverage (or/and train) performance for improving saliency metric scores, such as smoothing/thresholding (see Table~\ref{table:qualitative}, rows 4-5) or a center Gaussian (see Table~\ref{table:qualitative}, row 5). We should also consider that some of these models are already finetuned for synthetic images (e.g. SAM-ResNet). Our Model (Table~\ref{table:qualitative}, row 6), that has not been trained on any ground truth saliency data, has shown to be robust on these two distinct scenarios. It is also interesting to observe that our model can correctly detect multiple salient objects (see Table~\ref{table:qualitative}, columns 1,3-4).

% EXP 3: comparison SoA
Finally, we have compared the scores with classical hand-crafted saliency models (i.e. IKN, AIM, SDLF and GBVS) and state of the art deep saliency models (i.e. ML-Net, DeepGazeII, SAM and SalGAN) which are trained on ground truth fixation data. The results are summarized in Table~\ref{table:SOA_realImages} and Table~\ref{table:SOA_SyntheticsImages}.
On real images (TORONTO, MIT1003, and KTH) we perform similarly as the best hand-crafted method, and are only slightly outperformed by recent supervised deep learning methods (Table~\ref{table:SOA_realImages}). On synthetic images with synthetic and pop-out patterns (CAT2000, SID4VAM) we outperform all other deep saliency models (Table~\ref{table:SOA_SyntheticsImages}). On CAT2000 we even obtain the new state-of-the-art, whereas on SID4VAM we are second behind GBVS. This suggests that we are able to extract bottom-up attention maps but we are not biased to specific features of the dataset. Considering that our model is not trained on any fixation data, it is remarkable that our model can obtain competitive saliency estimation results. 

\paragraph{\textbf{Difference with Neural Networks for Saliency Detection}}

We compared our approach against other state-of-the-art deep saliency models, such as ML-Net \citep{mlnet2016}, SAM \citep{Cornia2018tip}, DeepGazeII \citep{Kummerer2016}, and SalGAN \citep{Pan2017SalGAN2}. The main difference between these models and our approach is that they use fixation data from saliency datasets in order to train a neural network. In other words, they require a saliency map as ground-truth. However, in our approach we do not require a saliency map as ground-truth. In contrast to the previous models, our approach is able to learn the saliency map as a side-effect when the network is trained end-to-end for an object recognition task. This difference is highlighted in Table~\ref{table:SOA_realImages} and Table~\ref{table:SOA_SyntheticsImages}, where the 'GT' column specifies which models require saliency maps as ground-truth for training. %The closest work to ours is \citep{murabito2018top}. However, their approach to generate saliency maps from RGB images, as a side-effect of a visual classification task, is supervised, which means they still require saliency maps to train the end-to-end network.}

%scores top-3 in AUC and SIM metrics for real image saliency datasets (TORONTO, MIT1003, KTH).

%Our model outperforms the hand-crafted saliency models in both AUC and SIM metrics (see Table~\ref{table:SOA} on real images,   \CF{exactly in the row: UCB%older}), and outperforms all other deep saliency models with synthetic and pop-out patterns (CAT2000, SID4VAM). 

\begin{table*}[!h]
\scriptsize

\centering
\caption{Ablation of fusion and normalization on all saliency datasets (computed with AlexNet). We show results for the AUC-Judd metric. Top-1 fusion methods are in \textbf{bold}.} 
\begin{tabular}{l|c|c|ccccc} 
 \hline\hline
   & DVA & Fusion & TORONTO & MIT1003 &  KTH & CAT2000 & SID4VAM \\ \hline
  \parbox[t]{2mm}{\multirow{6}{*}{\rotatebox[origin=c]{90}{Circular}}} 
         & 35 x 2 & Mult   & 0.616 & 0.595 & 0.521 & 0.620 & 0.525\\
    & & Sum    & 0.770 & 0.745 & 0.600 & 0.732 & 0.741\\
         & 35 x 5  & Mult   & 0.762 & 0.718 & 0.578 & 0.753 & 0.605\\
   &   &Sum   & 0.781 & 0.768 & 0.609 & 0.780 & 0.736\\
          & 35 x 14 & Mult  & 0.792 & 0.792 & 0.632 & 0.812 & 0.730\\
   &    &Sum  & 0.789 & 0.794 & 0.635 & 0.819 & 0.722\\
   \hline
  \parbox[t]{2mm}{\multirow{6}{*}{\rotatebox[origin=c]{90}{Ellipsoid}}} 
         & 35 x 2 & Mult   & 0.640 & 0.651 & 0.527 & 0.678 & 0.540\\
    & & Sum    & 0.776 & 0.758 & 0.597 & 0.751 & 0.740\\
         & 35 x 5  & Mult   & 0.780 & 0.724 & 0.581 & 0.759 & 0.611\\
   &   &Sum   & 0.788 & 0.771 & 0.620 & 0.790 & 0.740\\
          & 35 x 14 & Mult  & 0.800 & 0.799 & 0.639 & 0.812 & 0.730\\
   &    &Sum  & \textbf{0.801} & \textbf{0.800} & \textbf{0.640} & \textbf{0.820} & 0.730\\
   \hline
   SCB & -  & Mult  & 0.796 & 0.796 & 0.628 & 0.812 & \textbf{0.746}\\
   SCB & -  & Sum & 0.793 & 0.795 & 0.634 & 0.787 & 0.741\\
   \hline
\end{tabular}
  % \\\hspace{2mm}\small \CF{* The values were computed with AlexNet}

\label{table:fusion}

\end{table*}

\begin{table*}[!h]
\scriptsize
\centering
\caption{Qualitative results for Toronto, MIT1003 and SID4VAM. Rows provide results for different models. Results of \emph{Ours} are computed with ResNet152.} 
%\begin{tabular}{ c|ccc|ccc }
    \begin{tabular}{|c||cc||cc||cc|}

\hline
  & \textbf{TORONTO} &  & \textbf{MIT1003}   & & \textbf{SID4VAM} &\\
 \hline \hline
\begin{sideways} \centering Original Images\end{sideways}&
  \includegraphics[width=0.9in,height=0.55in]{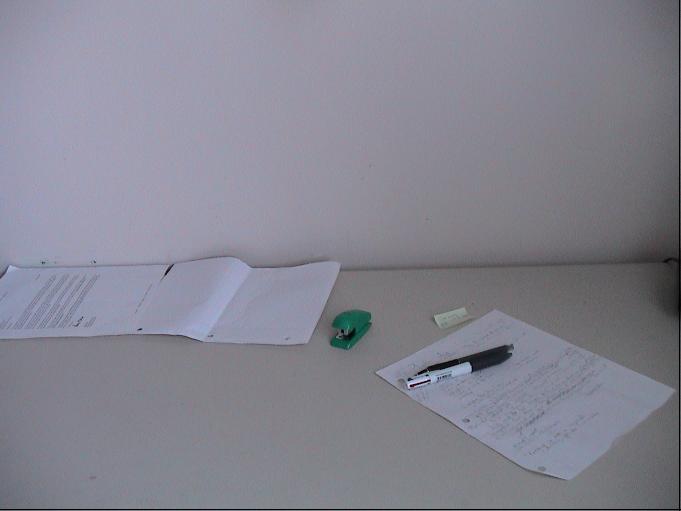} & 
  
  \includegraphics[width=0.9in,height=0.65in]{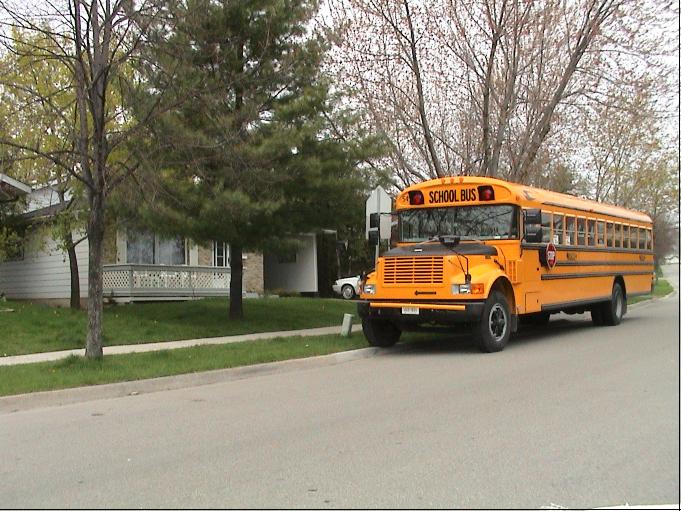} &
  
  %MIT1003
 \includegraphics[width=0.9in,height=0.55in]{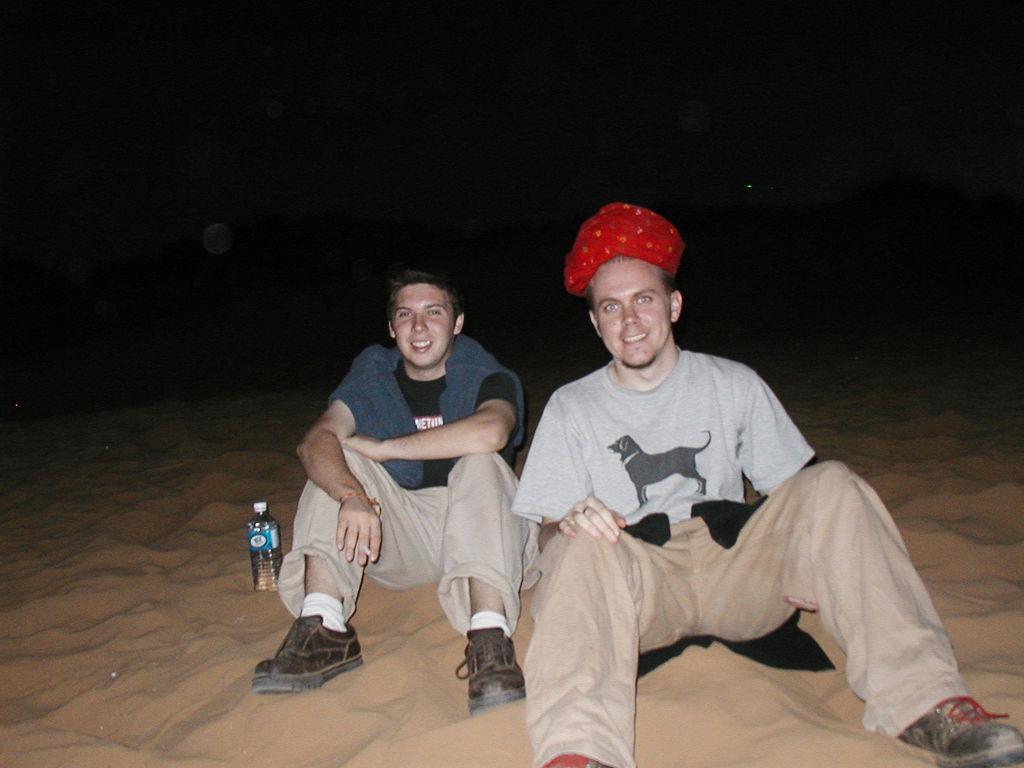} &

\includegraphics[width=0.9in,height=0.55in]{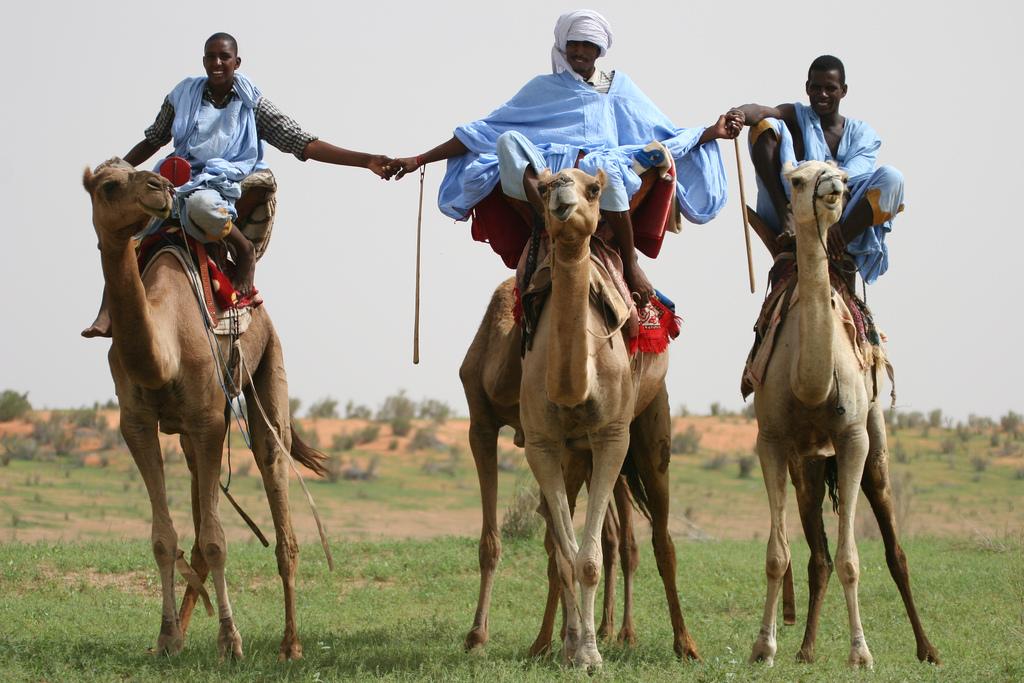} & 

% SID4VAM

\includegraphics[width=0.9in,height=0.55in]{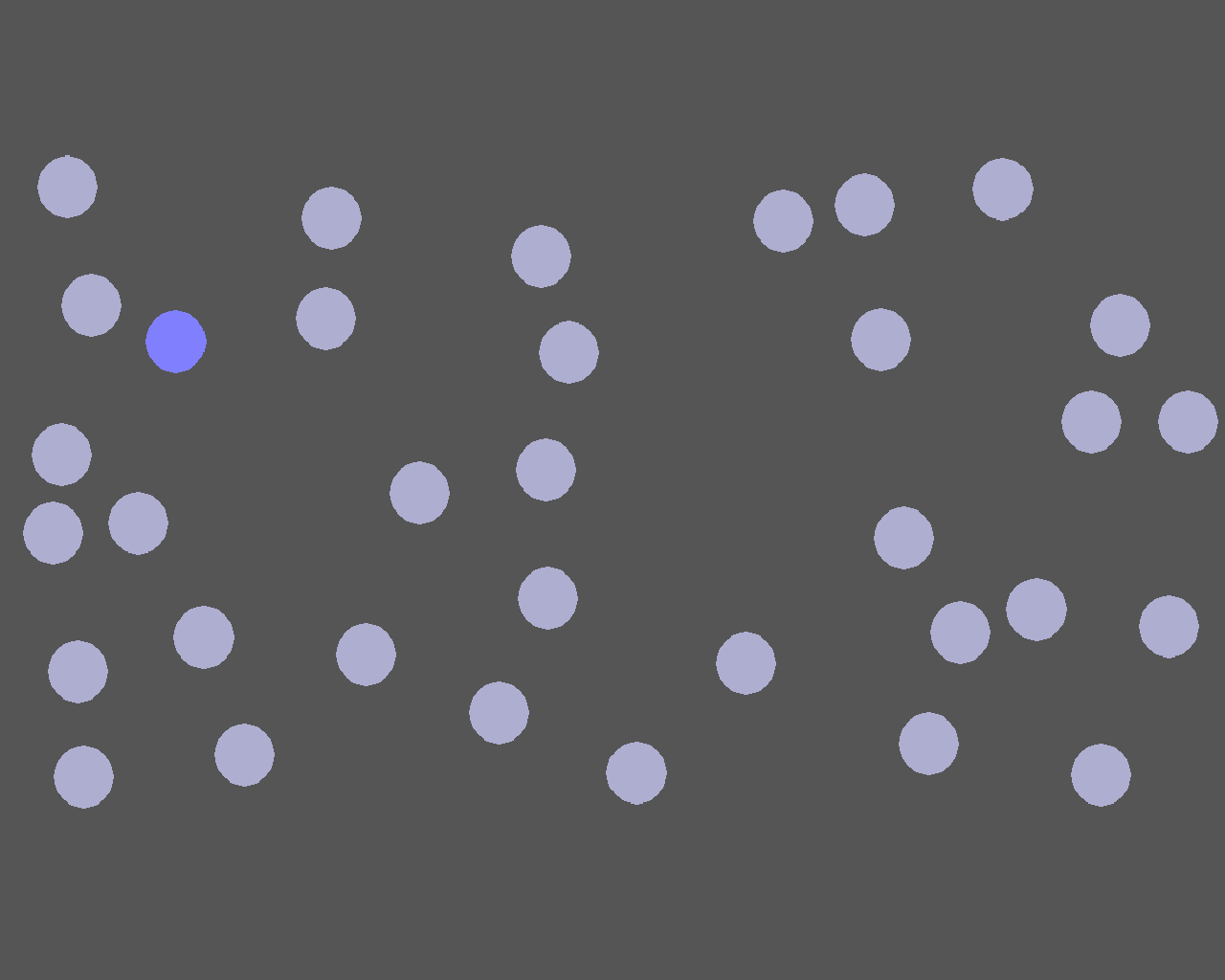} & 
\includegraphics[width=0.9in,height=0.55in]{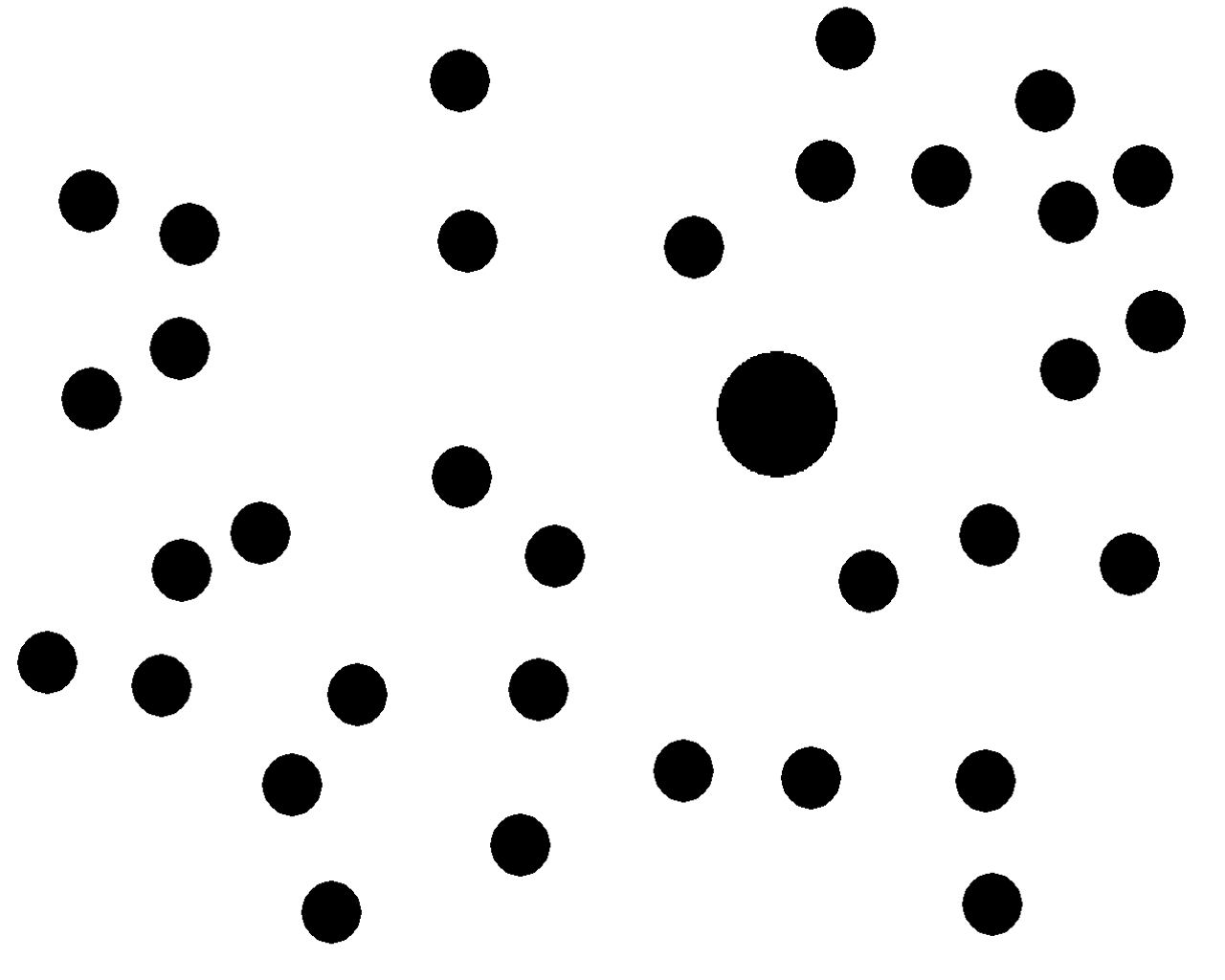} \\ 
\hline

%\begin{sideways} Human \end{sideways}
\begin{sideways} \centering Human \end{sideways}
&
  \includegraphics[width=0.9in,height=0.55in]{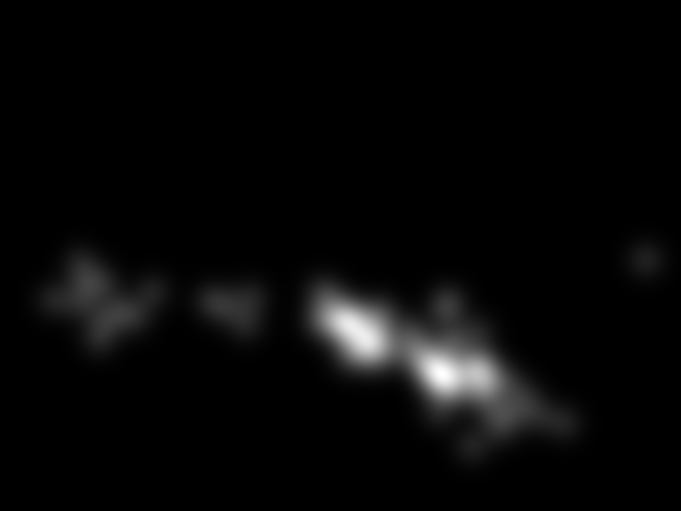} &   \includegraphics[width=0.9in,height=0.65in]{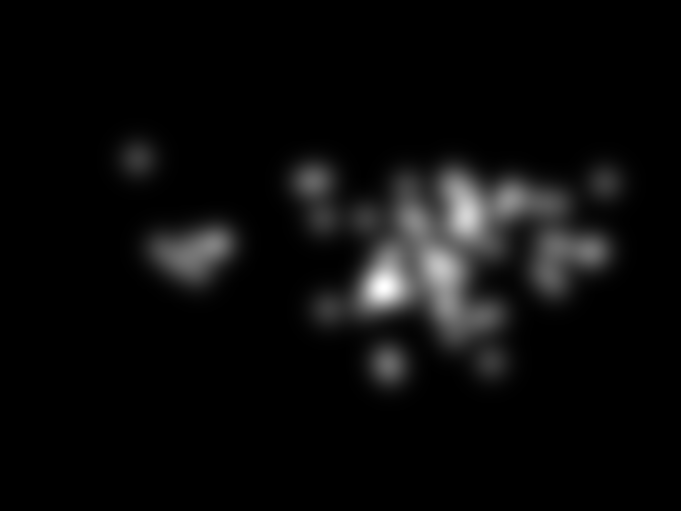} &

% MIT1003  
  
\includegraphics[width=0.9in,height=0.55in]{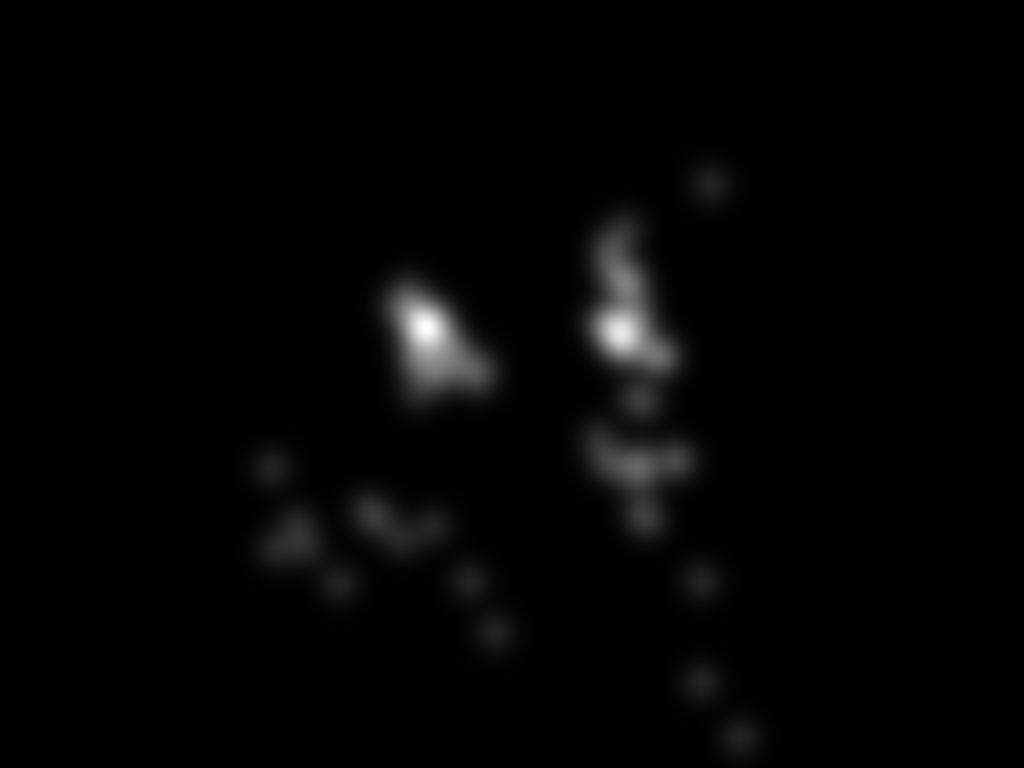} &

\includegraphics[width=0.9in,height=0.55in]{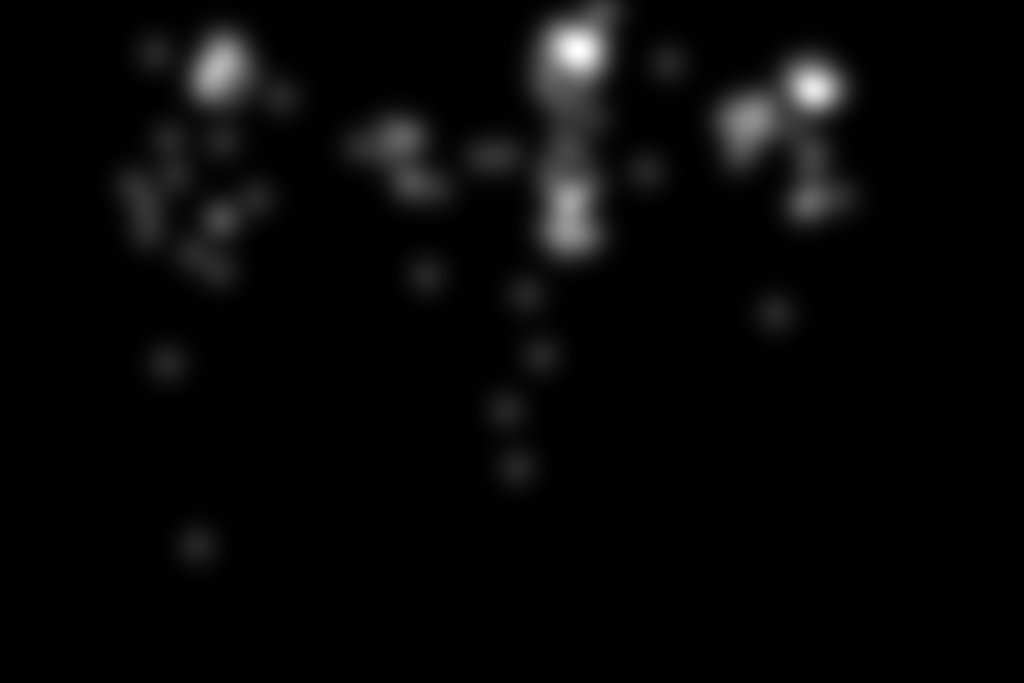} &  

% sid4vam
\includegraphics[width=0.9in,height=0.55in]{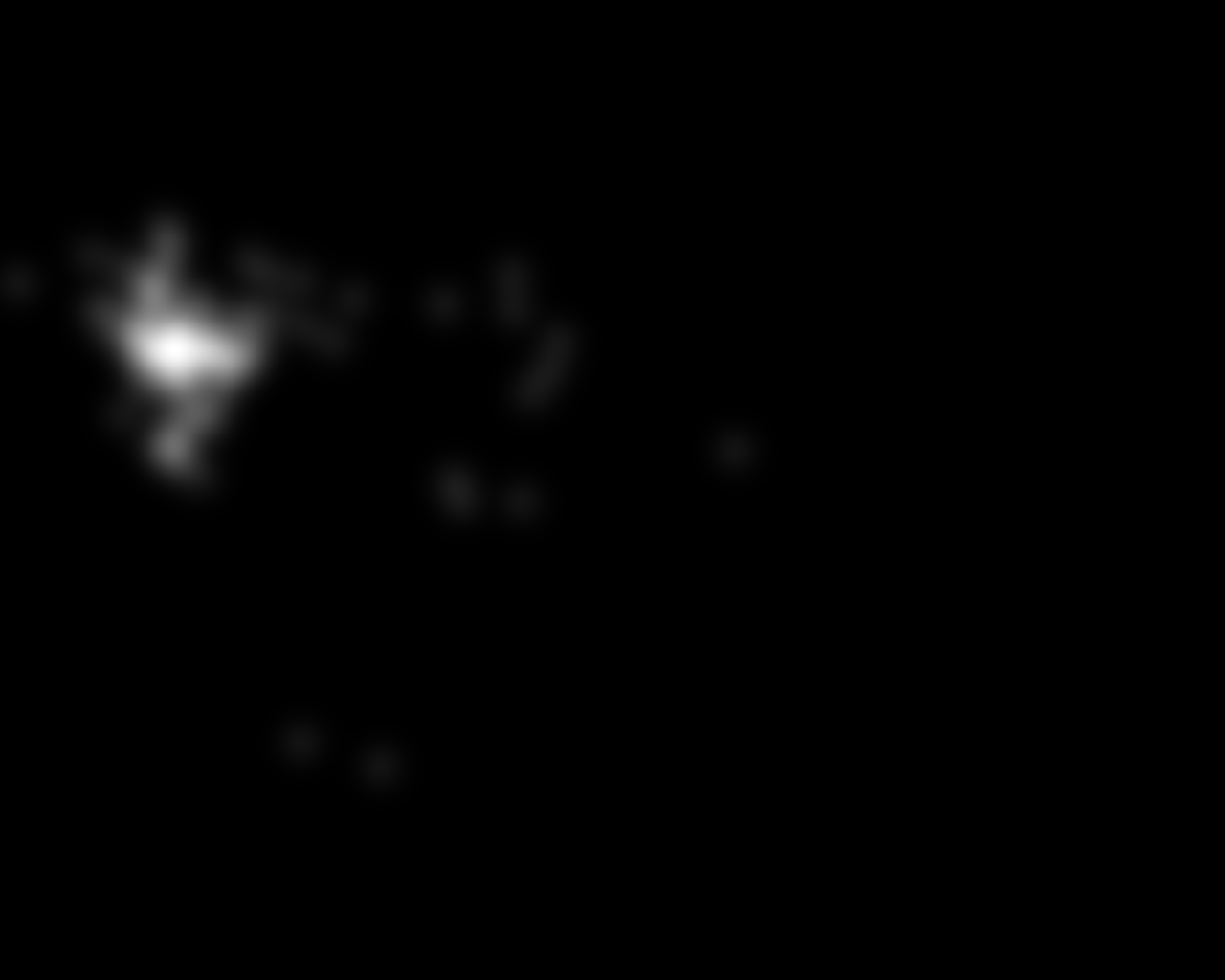} & 
\includegraphics[width=0.9in,height=0.55in]{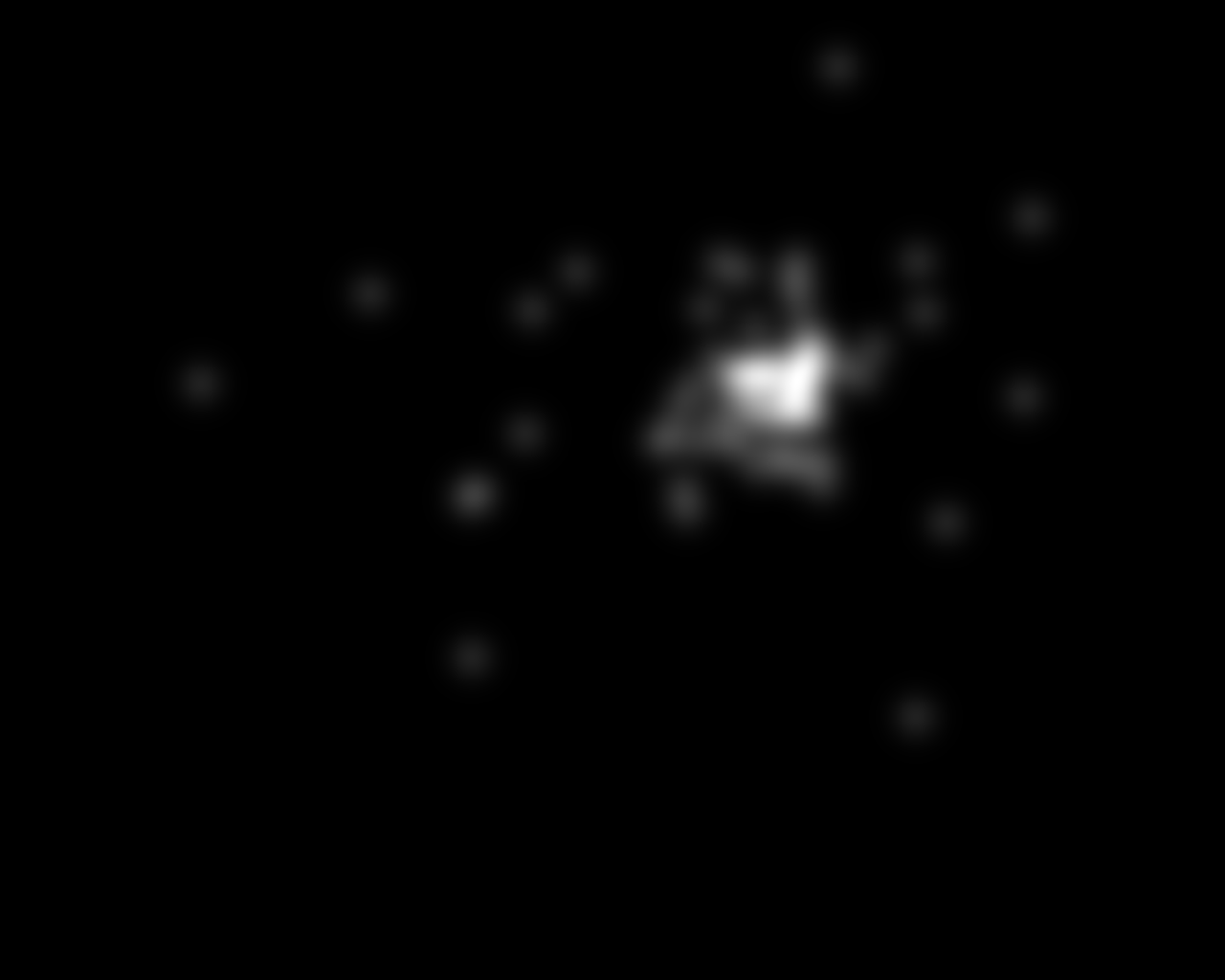} \\  
\hline

\begin{sideways} GBVS \end{sideways}
&
  \includegraphics[width=0.9in,height=0.55in]{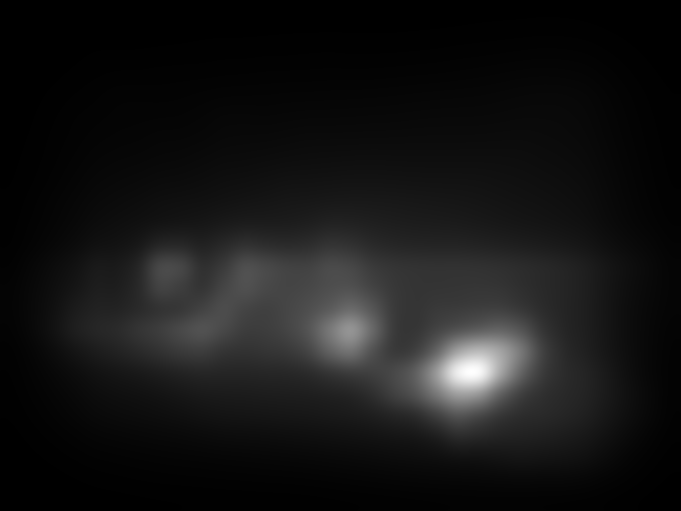} &  \includegraphics[width=0.9in,height=0.65in]{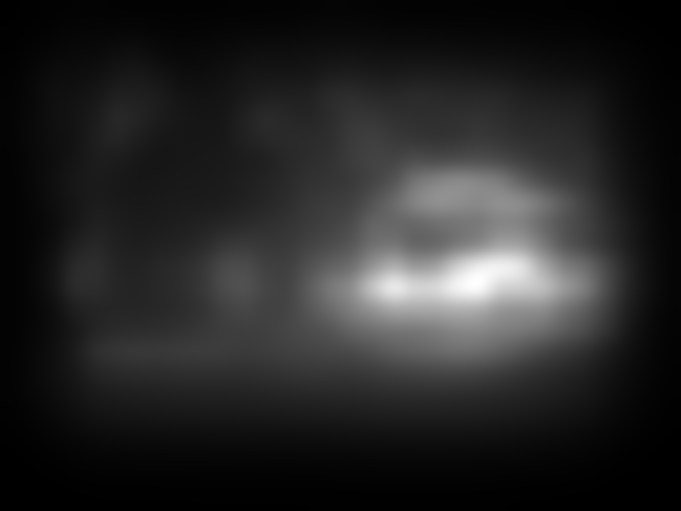} &
  
  % MIT
\includegraphics[width=0.9in,height=0.55in]{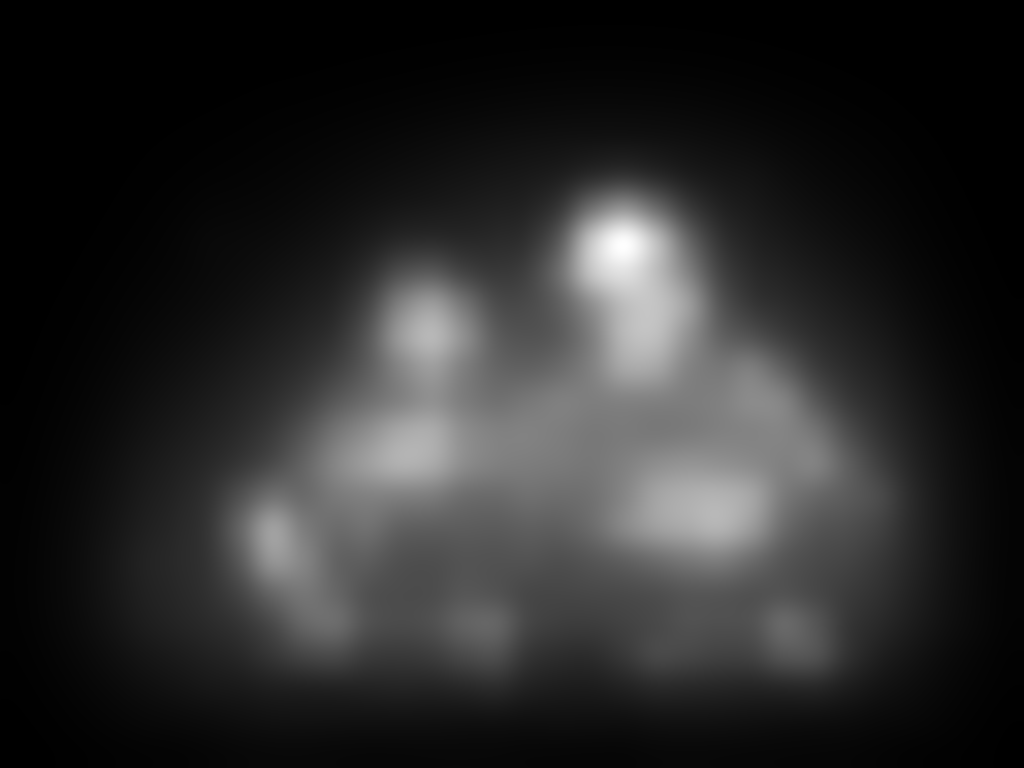} &

\includegraphics[width=0.9in,height=0.55in]{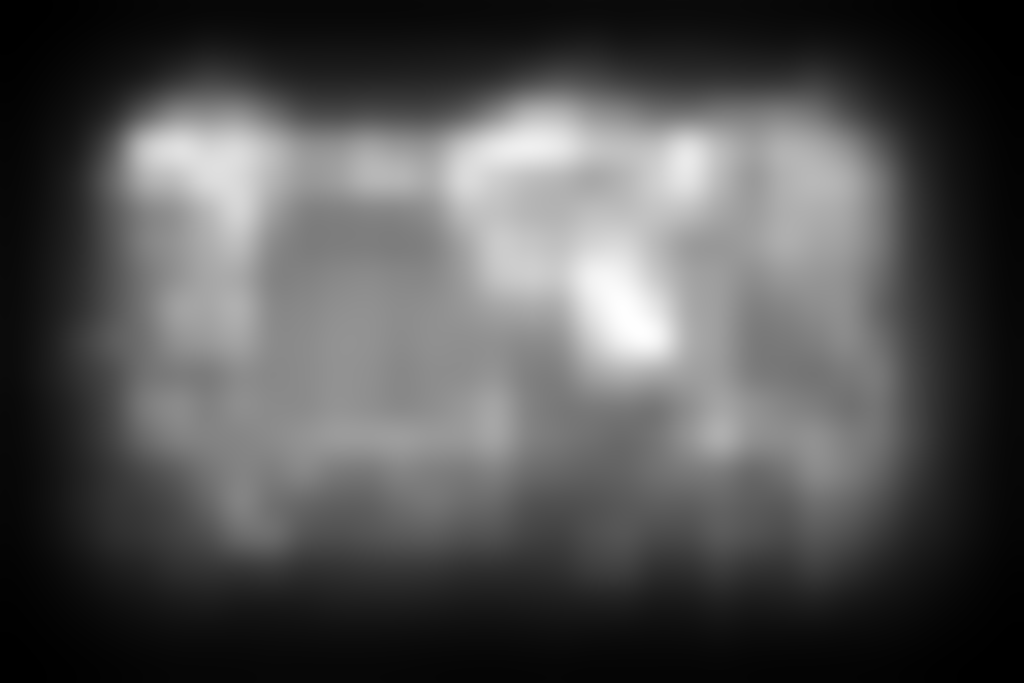} & 

% SID4VAM
\includegraphics[width=0.9in,height=0.55in]{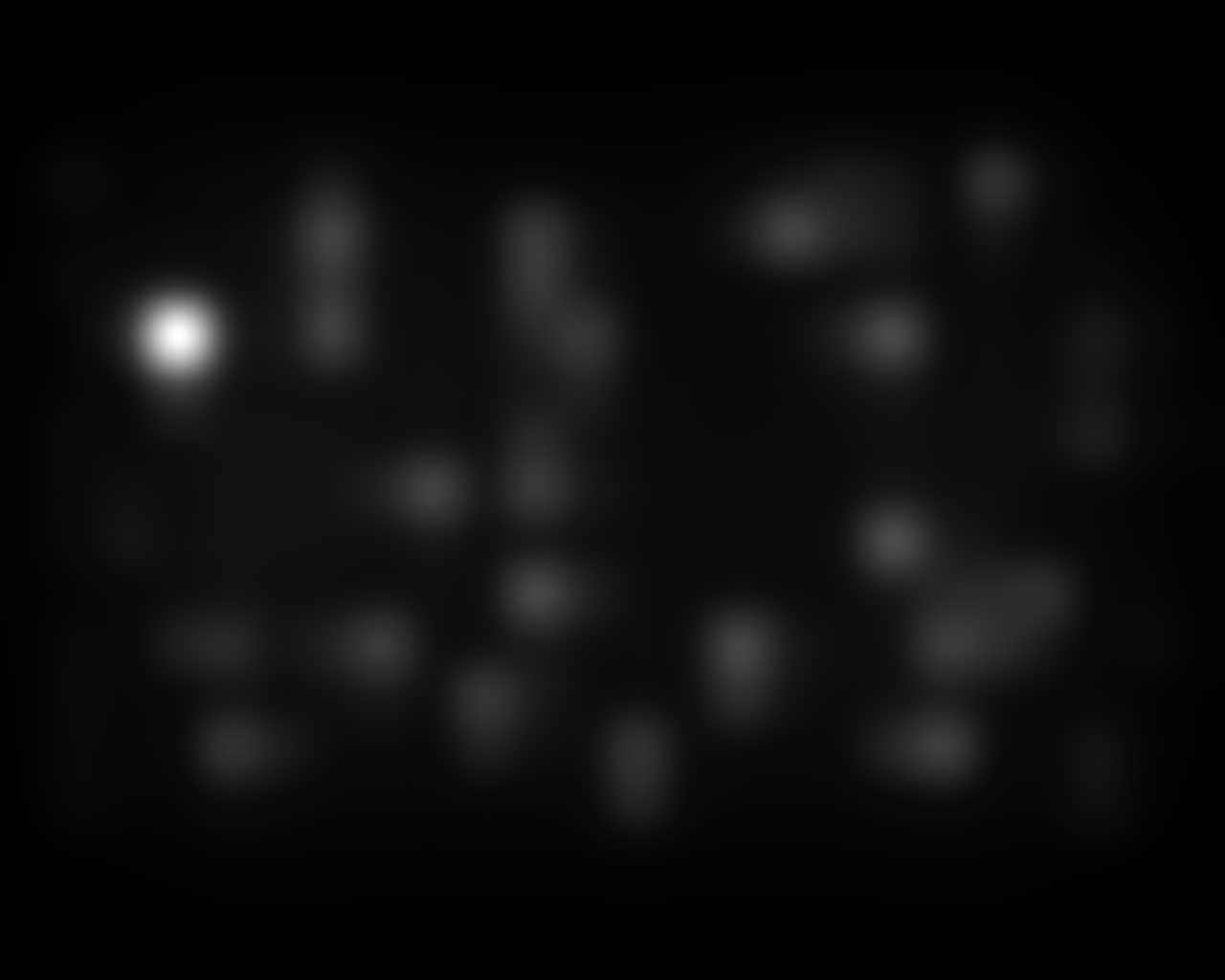} & 
\includegraphics[width=0.9in,height=0.55in]{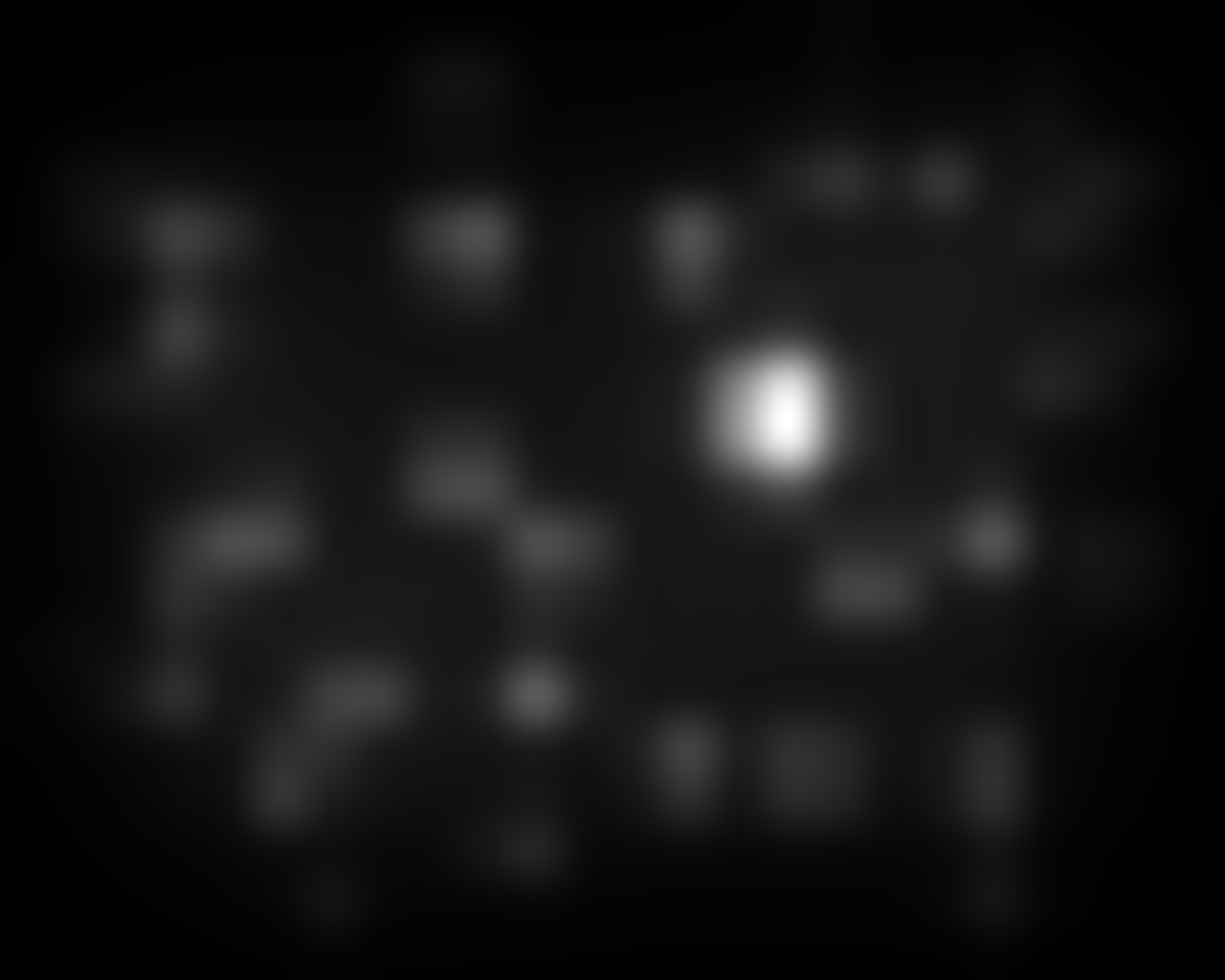} \\  
\hline

\begin{sideways} OpenSALICON \end{sideways}&
  \includegraphics[width=0.9in,height=0.55in]{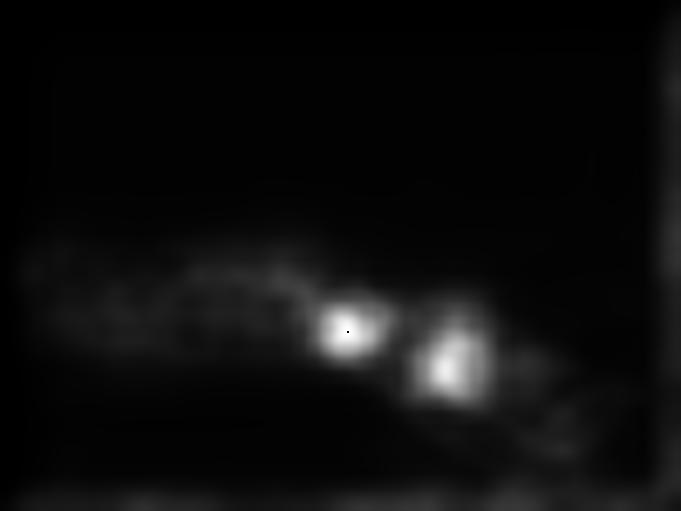} &  
  \includegraphics[width=0.9in,height=0.65in]{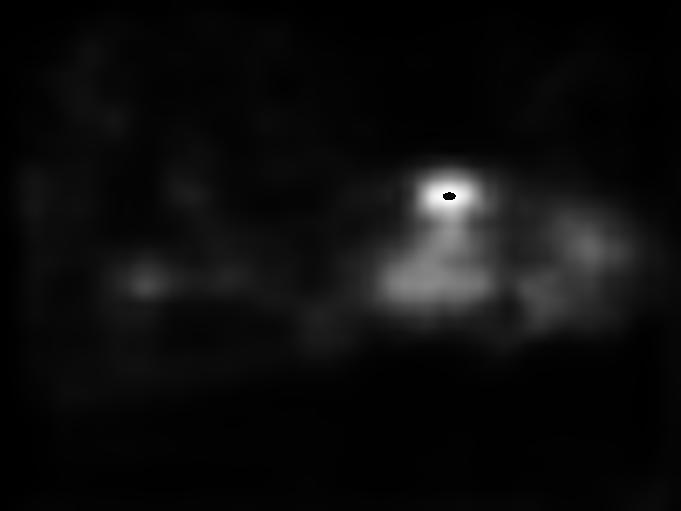} &
  
  % MIT1003
\includegraphics[width=0.9in,height=0.55in]{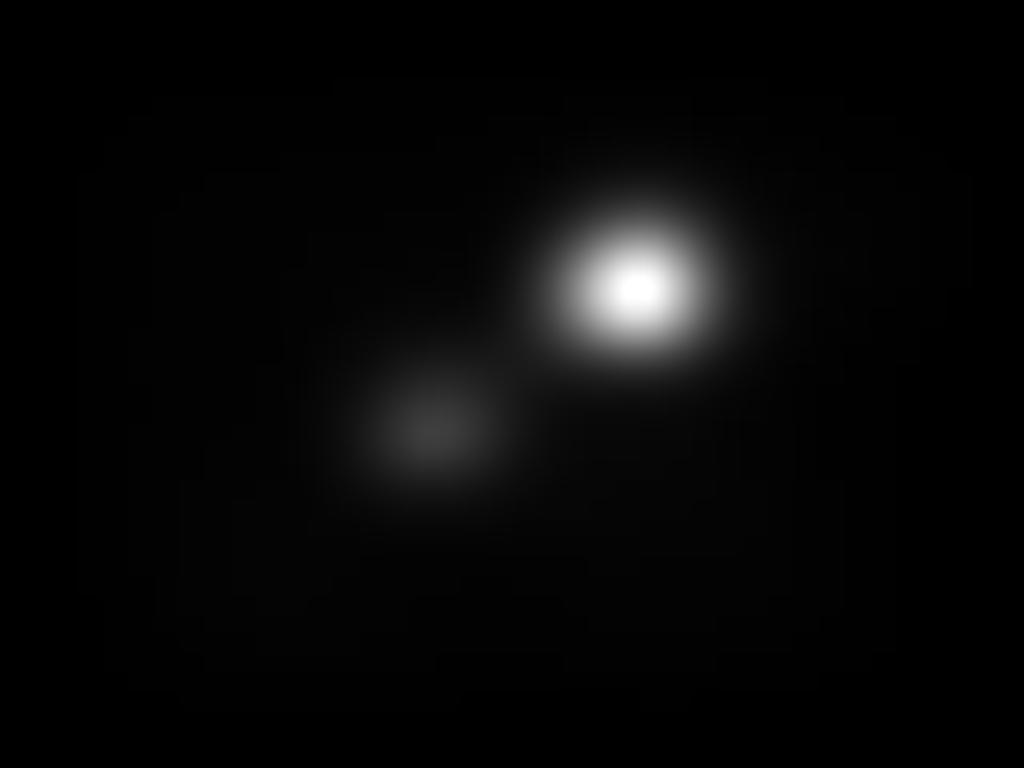} &

\includegraphics[width=0.9in,height=0.55in]{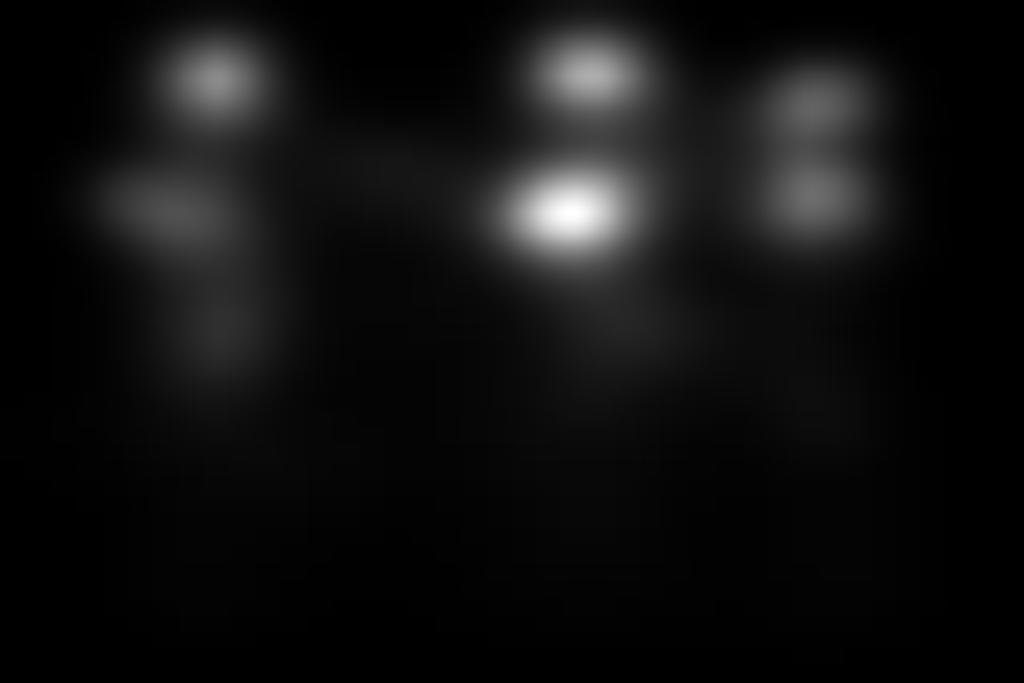} & 

% SID4VAM

\includegraphics[width=0.9in,height=0.55in]{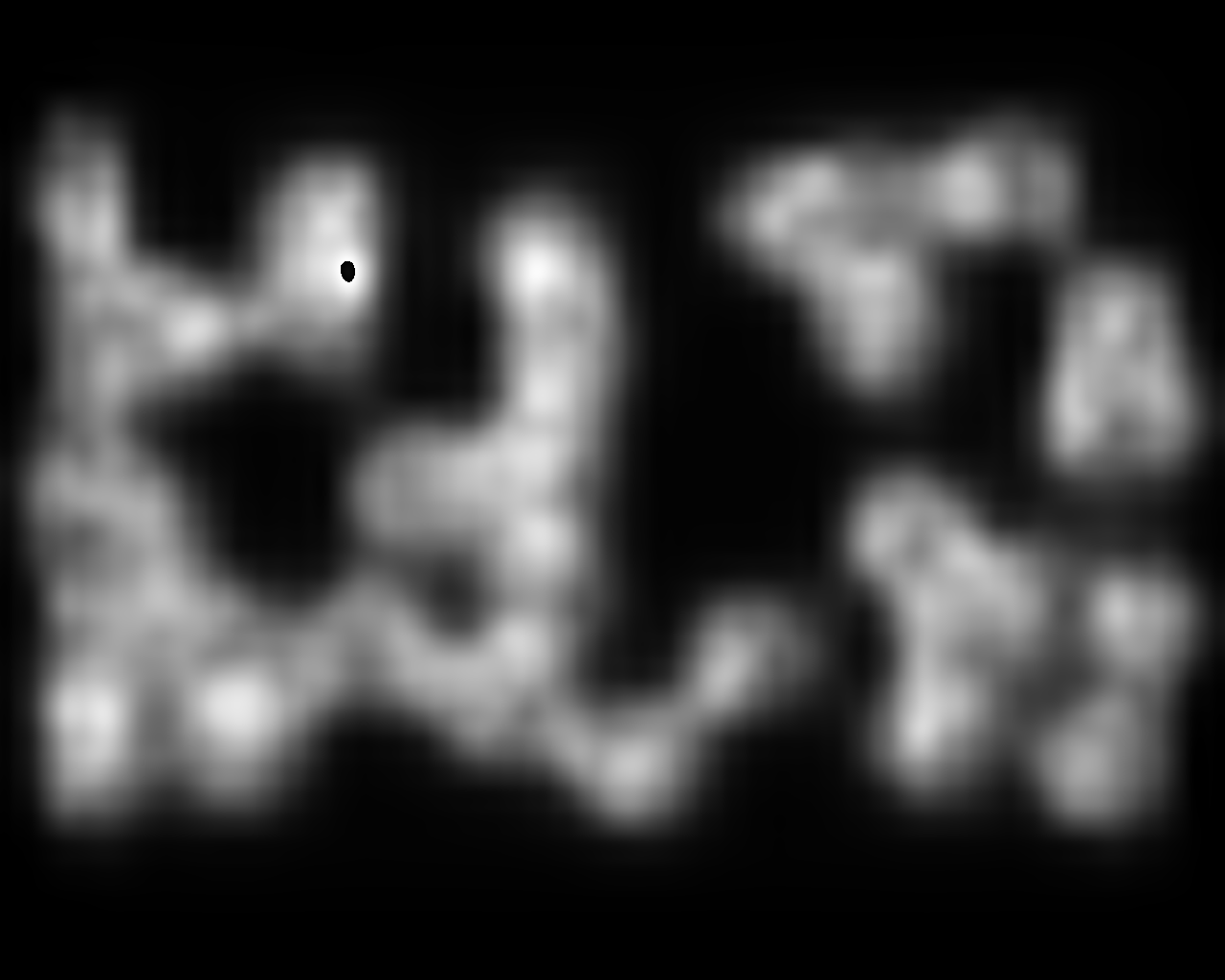} & 
\includegraphics[width=0.9in,height=0.55in]{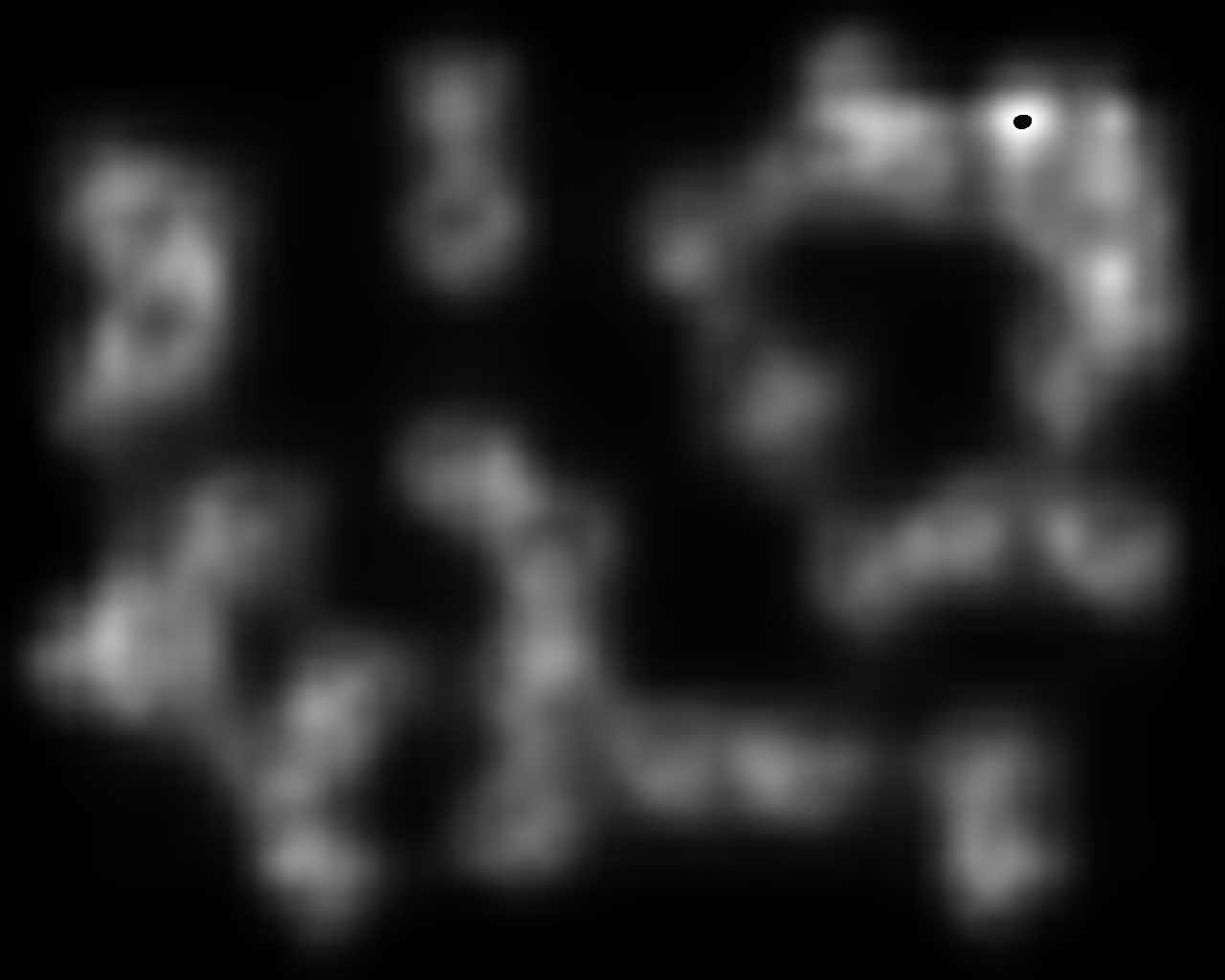} \\  
\hline

\begin{sideways} Sam-ResNet \end{sideways}&
  \includegraphics[width=0.9in,height=0.55in]{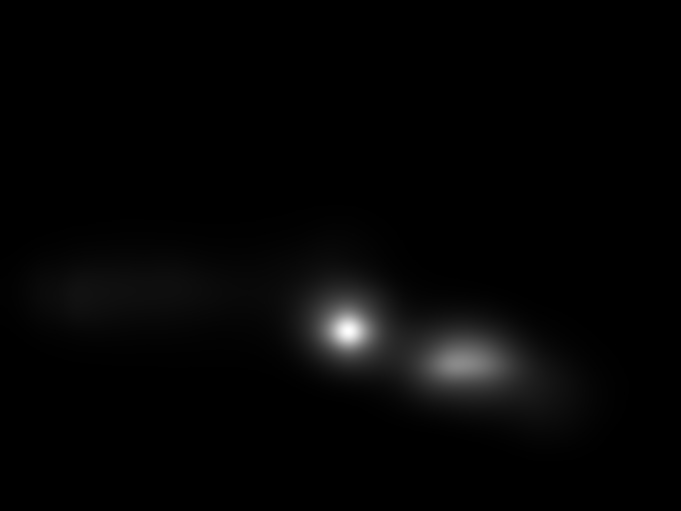} & 
   \includegraphics[width=0.9in,height=0.65in]{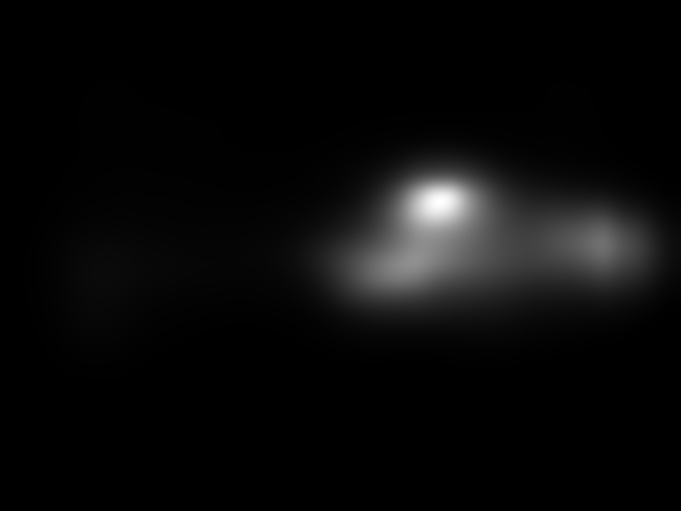} &
  
  % MIT1003
\includegraphics[width=0.9in,height=0.55in]{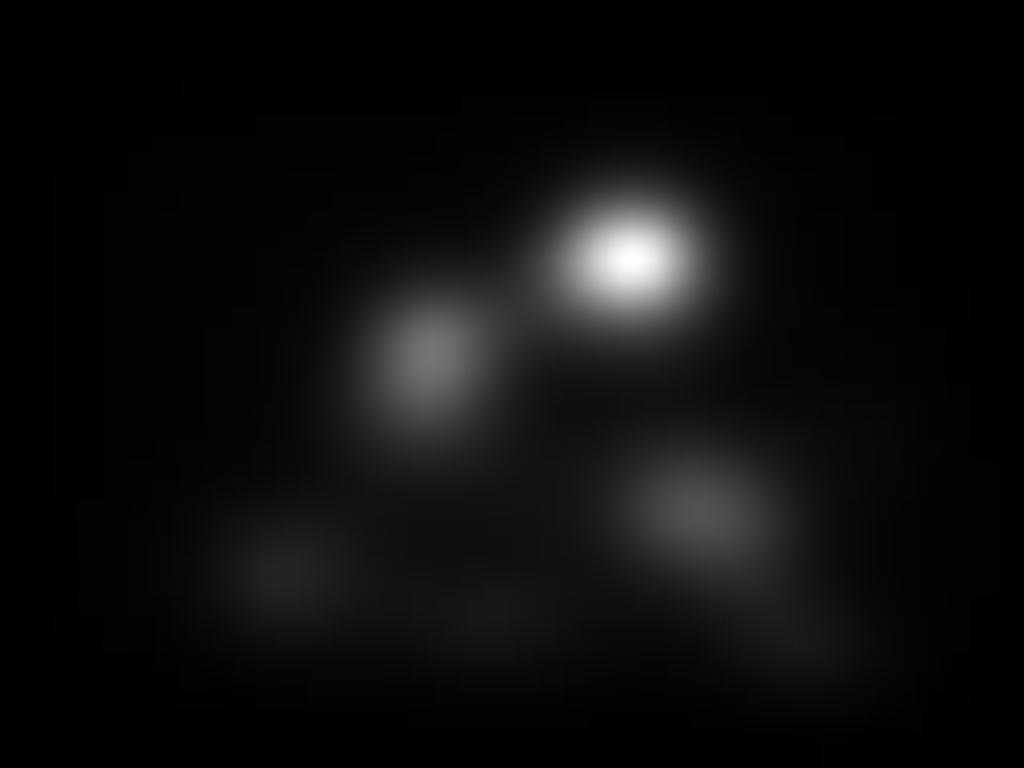} &

\includegraphics[width=0.9in,height=0.55in]{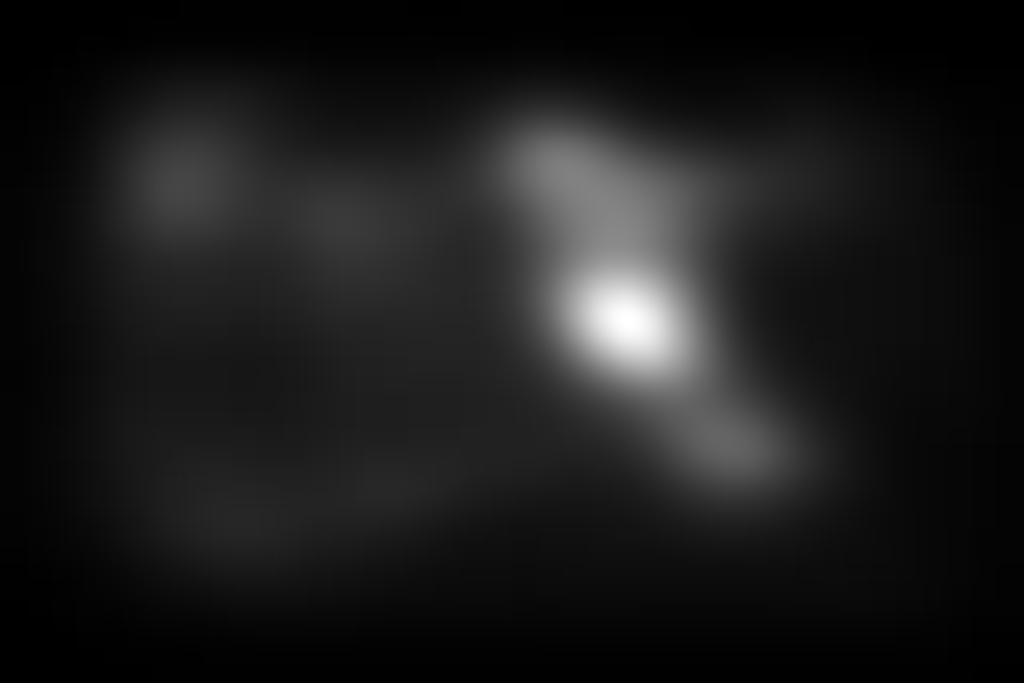} & 

% SID4VAM
\includegraphics[width=0.9in,height=0.55in]{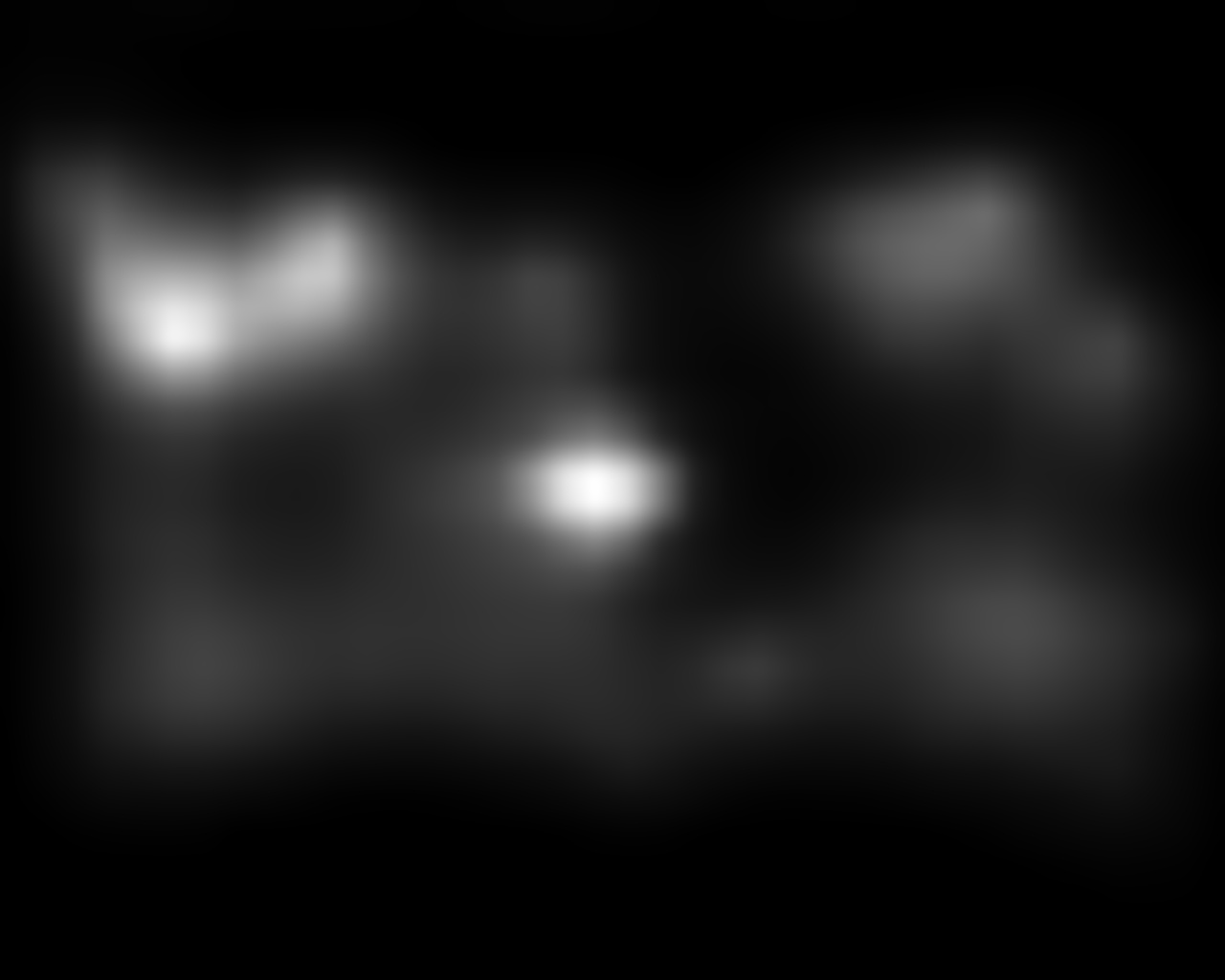} & 
\includegraphics[width=0.9in,height=0.55in]{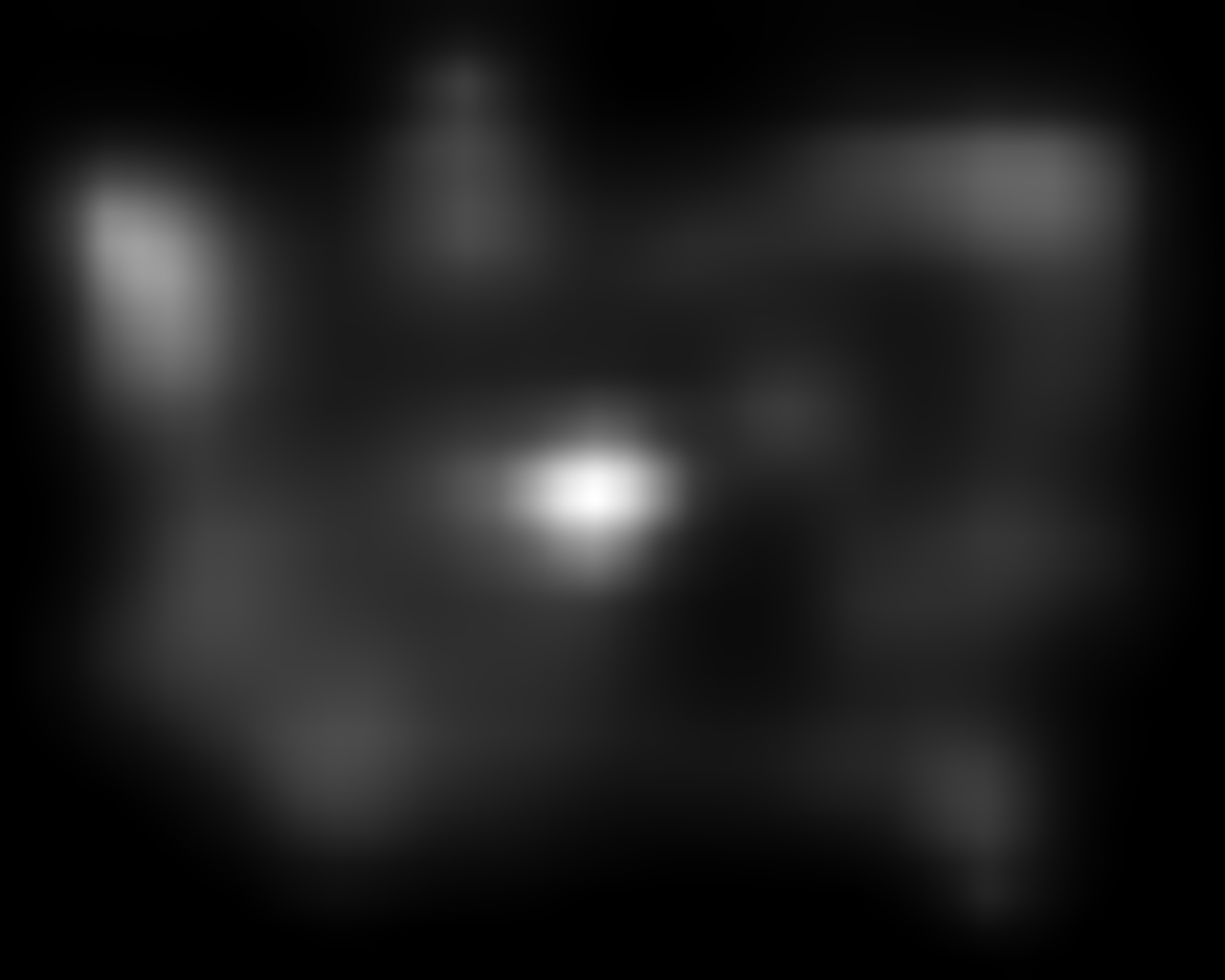} \\  
\hline

\begin{sideways} Ours \end{sideways}&
  \includegraphics[width=0.9in,height=0.55in]{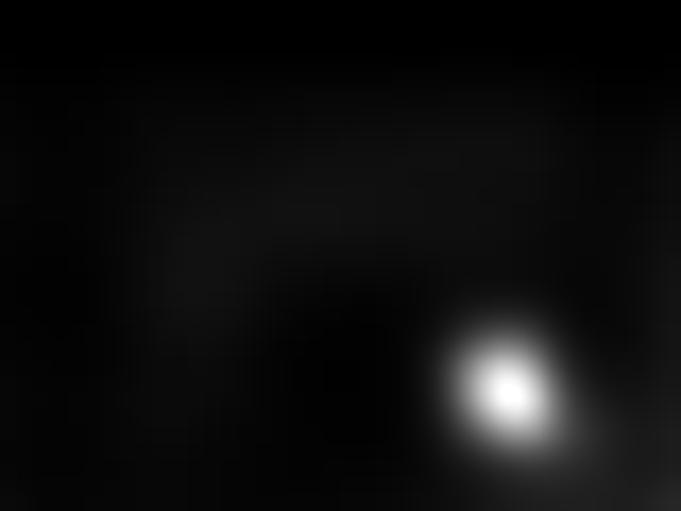} &  
  \includegraphics[width=0.9in,height=0.65in]{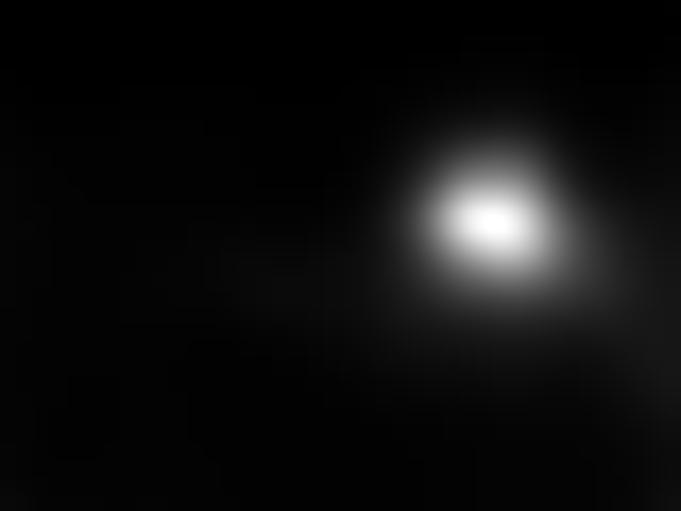}&
  
  % mit1003
 \includegraphics[width=0.9in,height=0.55in]{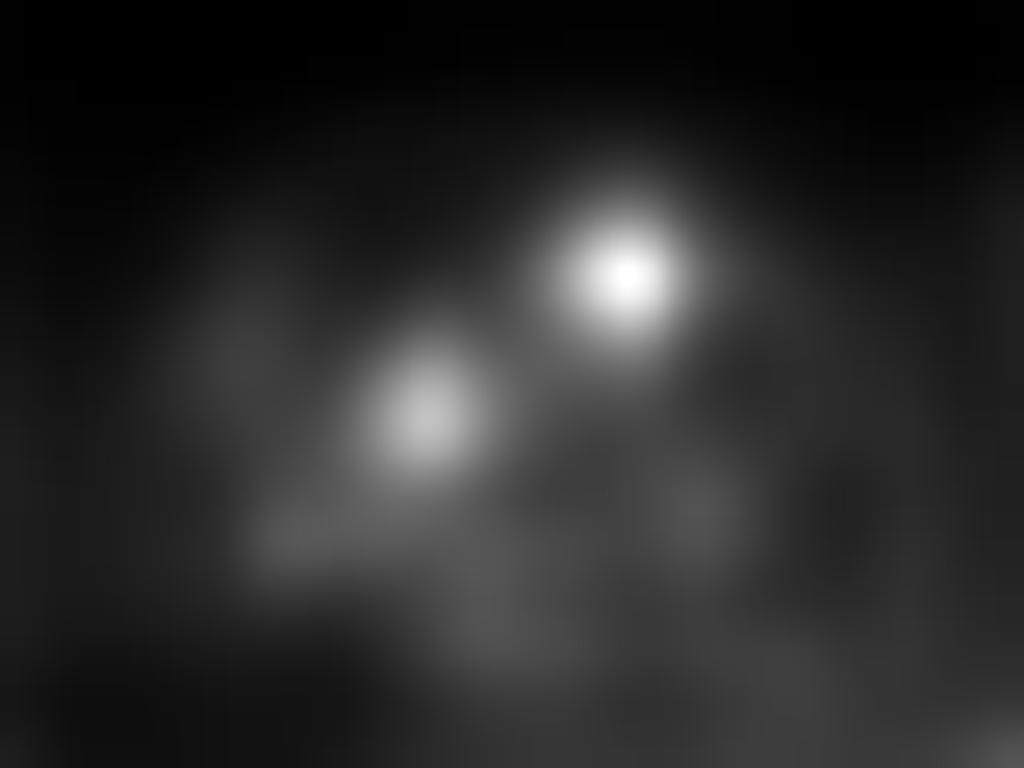} &

\includegraphics[width=0.9in,height=0.55in]{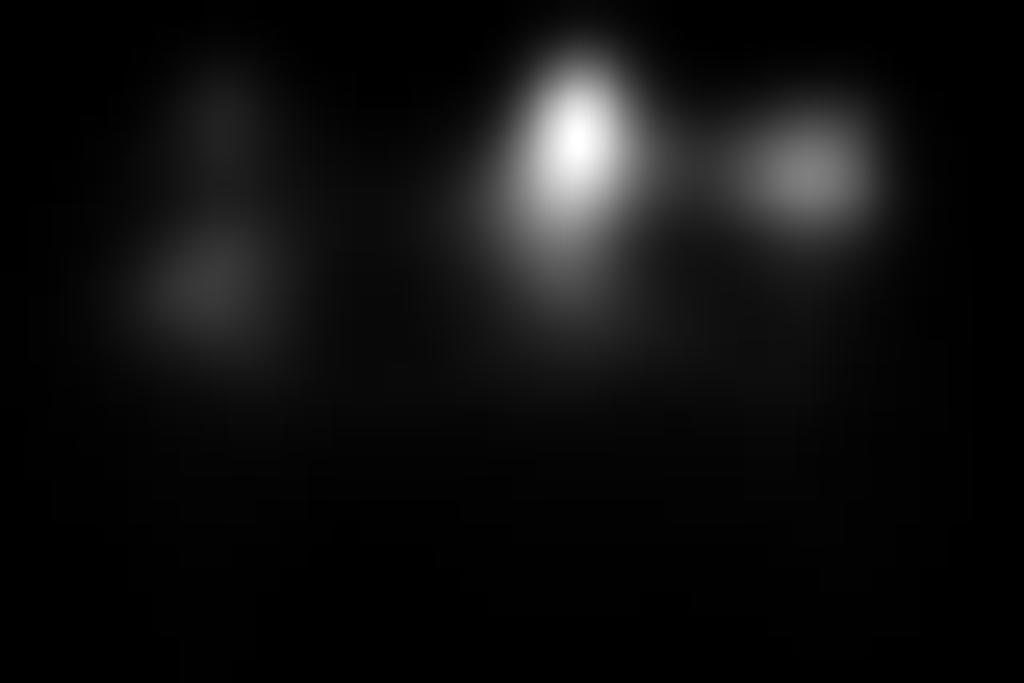} &  
% SID4VAM
\includegraphics[width=0.9in,height=0.55in]{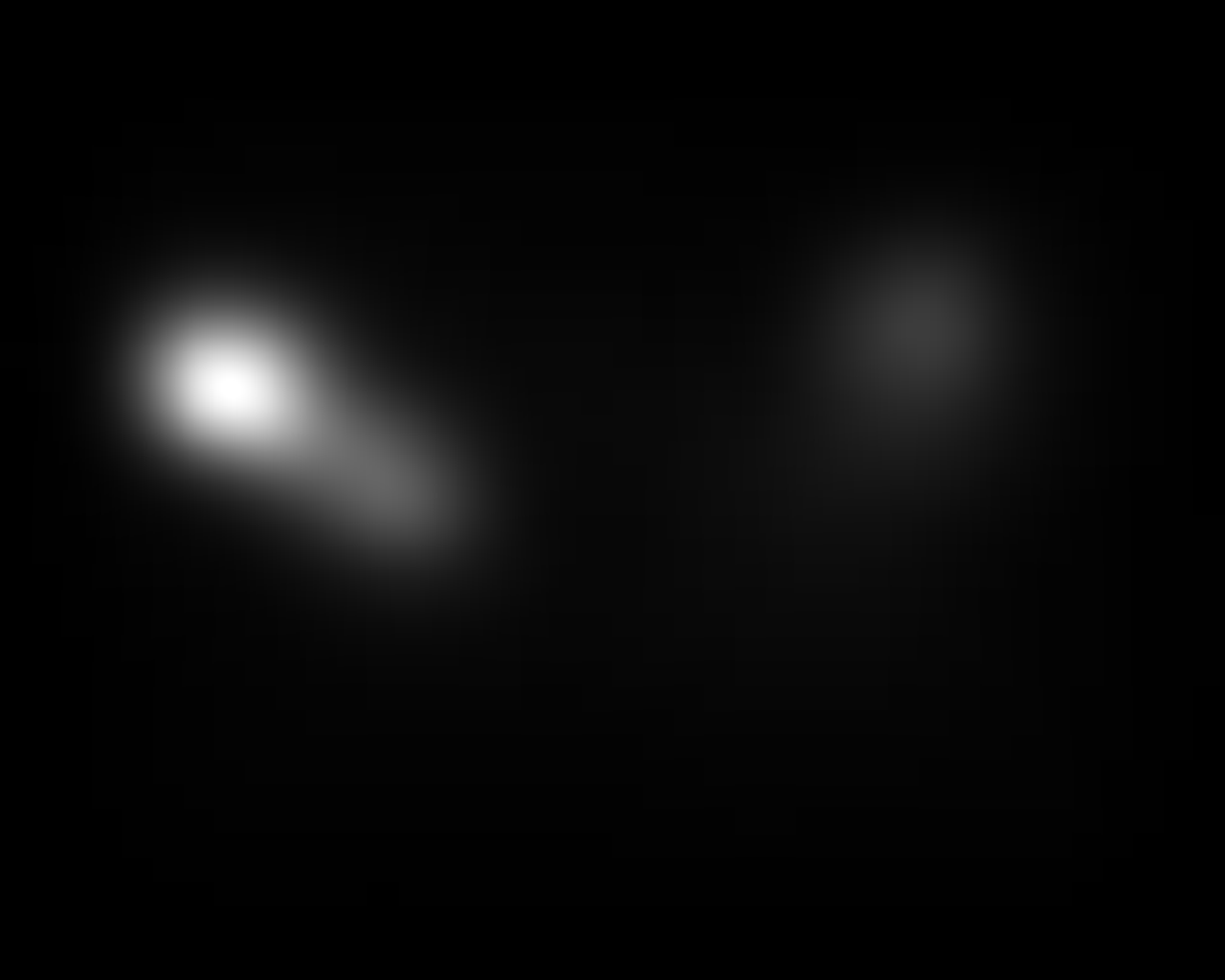} & 
\includegraphics[width=0.9in,height=0.55in]{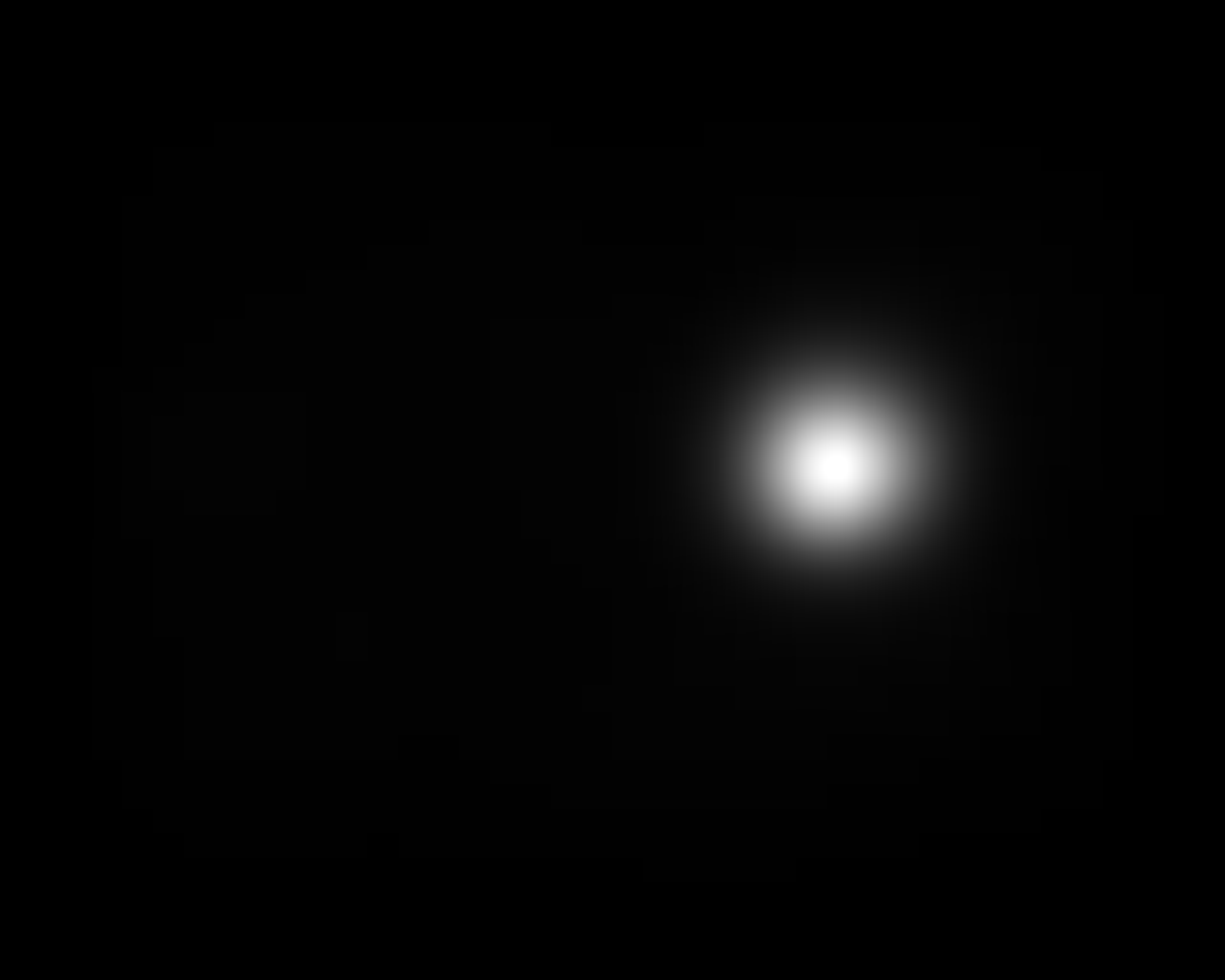} \\  
\hline

\hline
\end{tabular}
\label{table:qualitative}
\end{table*}

\begin{table*}[!h]
\centering
\caption{Benchmark on saliency metrics. We show results for the real Images datasets \textbf{Toronto (left) , MIT1003 (middle) and KTH (right)} metrics comparing to state of the art. We show which methods require ground truth (GT) saliency maps for training. Results of \emph{Ours} are computed with ResNet152. Top-1 methods are in \textbf{bold}.}
\resizebox{1\textwidth}{!}{

\begin{tabular}{|l|c||ccccc|} 
 \hline\hline
    \textbf{Dataset}  & \multicolumn{6}{c|}{Toronto}\\
\hline
    \textbf{Model} & \textbf{GT} & \textbf{AUC-Judd} & \textbf{CC} &  \textbf{NSS} & \textbf{SIM} & \textbf{sAUC }\\ \hline
     \textbf{IKN}  & \bm{$\times$} & 0.794 & 0.421 & 1.246 & 0.366 & 0.650 \\
   \textbf{AIM}    & \bm{$\times$}  & 0.727 & 0.292 & 0.883 & 0.314& 0.663\\
   \textbf{SDLF}   & \bm{$\times$} & 0.714 & 0.267 & 0.813 & 0.304 & 0.664\\
   \textbf{GBVS}   & \bm{$\times$} & 0.817 & 0.487 & 1.431 & 0.397& 0.632\\
  \hline
  \textbf{DeepGazeII}  & \bm{$\surd$} &\textbf{0.850} & 0.495 & 1.455 & 0.325 & 0.763\\
   \textbf{ML-Net}      & \bm{$\surd$} &0.845 & 0.598 & 1.903 & 0.489 & \textbf{0.684}\\
   \textbf{SAM-VGG}     & \bm{$\surd$} & 0.569 & 0.055 & 0.158 & 0.214 & 0.506\\
   \textbf{SAM-ResNet}  & \bm{$\surd$} &\textbf{0.850} & \textbf{0.612} & \textbf{1.955}  & \textbf{0.516}  & 0.666 \\  
  \textbf{SalGAN}      & \bm{$\surd$} &0.821 & 0.552 & 1.891  & 0.435 & 0.715\\
  \hline

  \textbf{Ours}          & \bm{$\times$} &0.782 & 0.538 & 1.643 & 0.439  & 0.641 \\
  \textbf{Ours + UCB}   & \bm{$\times$} &0.813 & 0.448 & 1.252 & 0.449& 0.567\\
  \textbf{Ours + SCB}   & \bm{$\times$} &0.810 & 0.444 & 1.238 & 0.442 & 0.560 \\

   \hline
   \textbf{Humans (GT)} & \bm{$\times$} & 0.969 & 1.000 & 3.831 & 1.000 & 0.902\\  \hline   
   \hline
\end{tabular}
\begin{tabular}{|ccccc|} 
 \hline\hline
    \multicolumn{5}{|c|}{MIT1003}  \\
\hline
 \textbf{AUC-Judd} & \textbf{CC} &  \textbf{NSS} & \textbf{SIM} & \textbf{sAUC } \\ \hline
     0.760  & 0.305 & 1.019  & 0.290 & 0.636\\
   0.706  & 0.227 & 0.779 & 0.251  & 0.639 \\
   0.697 & 0.216 &  0.740 & 0.251  & 0.637\\
   0.807 & 0.374 & 1.246  & 0.324  & 0.621\\
  \hline
  0.849  & 0.432 & 1.482 & 0.360 & \textbf{0.773}\\
    0.839  & 0.535 & 1.918 & 0.424 & 0.695\\
   0.559 & 0.036 &  0.120 & 0.182 & 0.498\\
  0.854  & \textbf{0.579} & \textbf{2.079}  & \textbf{0.472} & 0.678\\  
    \textbf{0.856} & 0.552 & 1.891 & 0.435 & 0.715\\ 
  \hline

  0.723 & 0.253 & 0.855 & 0.284 & 0.552\\
    0.810 & 0.360 & 1.170 & 0.307 & 0.551\\
       0.808 & 0.360 & 1.168 & 0.299 & 0.550\\

   \hline
  0.978 & 1.000 & 4.497 & 1.000 & 0.937\\  \hline   
   \hline
\end{tabular}

\begin{tabular}{|ccccc|} 
 \hline\hline
      \multicolumn{5}{|c|}{KTH} \\
\hline

 \textbf{AUC-Judd} & \textbf{CC} &  \textbf{NSS} & \textbf{SIM} & \textbf{sAUC }\\ \hline
     0.617 & 0.274 & 0.403 & 0.547 &  0.551\\
   0.572 & 0.179 & 0.274 & 0.523 & 0.552\\
    0.555 & 0.132 & 0.203 & 0.512 & 0.544\\
   0.649 & 0.351 & 0.505 & \textbf{0.563} & 0.532\\
  \hline
    0.648 & 0.348 & 0.530  & 0.549 & \textbf{0.597}\\
   0.658 & 0.384 & 0.579 & 0.557 & 0.568 \\
    0.525 & 0.058 & 0.074 & 0.354 & 0.501\\
    \textbf{0.660} & 0.371 & 0.570 & 0.508 & 0.548\\ 
    0.655 & \textbf{0.391} & \textbf{0.581} & 0.544 & 0.560\\
  \hline

   0.615  & 0.294 & 0.444  &  0.499 & 0.499\\
   0.645 & 0.327 & 0.468 & 0.505 & 0.517\\
  0.641 & 0.328  & 0.468 & 0.501 & 0.514\\

   \hline
   0.902  & 1.000 & 2.038 & 1.000 & 0.822\\ \hline   
   \hline
\end{tabular}
}
\label{table:SOA_realImages}
\end{table*}

\begin{table*}[!h]
\centering
\caption{Benchmark on saliency metrics. We show results for the Synthetic Images datasets \textbf{CAT2000 (left)  and SID4VAM (right)} metrics comparing to state of the art. We show which methods require ground truth (GT) saliency maps for training Results of \emph{Ours} are computed with ResNet152. Top-1 methods are in \textbf{bold}.}
\resizebox{0.8\textwidth}{!}{

\begin{tabular}{|l|c||ccccc|} 
 \hline\hline
  \textbf{Dataset}  & \multicolumn{6}{c|}{CAT2000}  \\
\hline
   \textbf{Model} & \textbf{GT} & \textbf{AUC-Judd} & \textbf{CC} &  \textbf{NSS} & \textbf{SIM} & \textbf{sAUC }\\ \hline
    \textbf{IKN}   &  \bm{$\times$} & 0.701 & 0.323 & 0.829 & 0.382 & 0.562\\
   \textbf{AIM}     & \bm{$\times$} &0.570 & 0.118  & 0.332 & 0.301 & 0.544\\
   \textbf{SDLF}    & \bm{$\times$} &0.573 & 0.111 & 0.308  & 0.309 & 0.550\\
   \textbf{GBVS}    & \bm{$\times$}  & 0.759 & 0.399 & 1.056 & 0.430 & 0.561\\
  \hline
   \textbf{DeepGazeII}   &  \bm{$\surd$}  & 0.612 & 0.174 & 0.480 & 0.335 & \textbf{0.571}\\
   \textbf{ML-Net}     & \bm{$\surd$} & 0.678 & 0.268 & 0.724 & 0.375 & 0.555\\
   \textbf{SAM-VGG}      & \bm{$\surd$}  & 0.625 & 0.123  & 0.320 & 0.322 & 0.508\\
   \textbf{SAM-ResNet}   &  \bm{$\surd$}  &0.766 & 0.518 & 1.356 & 0.456 & 0.546\\  
   \textbf{SalGAN}       &  \bm{$\surd$} &0.751 & 0.417 & 1.080 & 0.553 & 0.553\\
  \hline

  \textbf{Ours}           & \bm{$\times$}  & 0.722 & 0.310 & 0.855 & 0.406 &0.525 \\
  \textbf{Ours + UCB}     & \bm{$\times$}  & \textbf{0.822} & \textbf{0.610} & \textbf{1.574} & 0.544 & 0.531\\
  \textbf{Ours + SCB}     &  \bm{$\times$} & 0.820 & 0.607 & 1.566 & \textbf{0.561} & 0.530\\

   \hline
   \textbf{Humans (GT)}  &  \bm{$\times$}  &0.895 & 0.890 & 2.335 & 1.000& 0.623\\ \hline   
   \hline
\end{tabular}
\begin{tabular}{|ccccc|} 
 \hline\hline
 \multicolumn{5}{|c|}{SID4VAM}  \\
\hline
   \textbf{AUC-Judd} & \textbf{CC} &  \textbf{NSS} & \textbf{SIM} & \textbf{sAUC }\\ \hline
       0.686 & 0.283 & 0.878 & 0.380 & 0.608 \\
    0.570 & 0.122 & 0.473 & 0.224 & 0.557 \\
    0.620 & 0.156 & 0.585 & 0.322 & 0.596\\
    \textbf{0.747} & \textbf{0.400} & \textbf{1.464}  & \textbf{0.413} & \textbf{0.628}\\
  \hline
   0.612 & 0.174 & 0.480 & 0.335 & 0.571 \\
   0.700 & 0.283 & 0.883 & 0.373 & 0.595\\
   0.537  & 0.026 & 0.070 & 0.216 & 0.503\\
    0.727 & 0.305 & 0.967  & 0.388 & 0.600\\  
   0.715 & 0.287 & 0.883 & 0.373 & 0.593\\
  \hline

 0.699 & 0.301  & 1.114 & 0.379 & 0.598\\
   0.710 & 0.341 & 1.219 & 0.394 & 0.605\\
   0.711 & 0.339 & 1.216  & 0.388 & 0.601\\
 
   \hline
   0.943 & 1.000 & 4.204 & 1.000&  0.860\\  \hline   
   \hline
\end{tabular}

}

\label{table:SOA_SyntheticsImages}

\end{table*}

\section{Conclusions}

This study shows that saliency might be an intrinsic effect in image representation learning, and this can be obtained by training other tasks such as image classification. By training on ImageNet for image classification, we are able to extract saliency maps without the need of any ground truth saliency data. Our model obtains good results with various metrics and datasets, acquiring similar results to the state of the art, however, without the need of any ground truth saliency maps. We have added a study of which networks and typologies of center biases can affect saliency prediction. 

Our work is the first to show that saliency estimation can be derived as a side-effect of training an end-to-end deep neural network for object recognition. Interestingly, by optimizing to perform optimal object recognition, the network learns to put attention on locations which are considered to be salient for humans. Of special interest is the fact, that our saliency branch trained for the tasks of object recognition on ImageNet, a dataset with real images, obtains excellent results on synthetic saliency datasets which have very different characteristics. Possible improvements of our method could include finetuning with fixation data, enabling to tune the saliency branch (and/or the center bias) by training on some small selection of real fixation data. Also, saliency branches could be derived from other computer vision tasks, such as robots navigating through an environment or self-driving cars \citep{Wang2020iros}.

%Our results showed that scores vary considerately depending on dataset or network, as every dataset has specific set of features and parametization of experimental systematic tendencies, suggesting that there cannot be a unique solution for modeling the center bias in combination with saliency.

\section*{Acknowledgments}
The authors acknowledge the Spanish  project  PID2019-104174GB-I00 (MINECO, Spain). and the CERCA Programme of Generalitat de Catalunya. Carola Figueroa is supported by a Ph.D. scholarship from CONICYT, Chile.

\bibliographystyle{model2-names}
\bibliography{refs}

\end{document}